\documentclass[10pt]{article} % For LaTeX2e
\usepackage[preprint]{tmlr}
\usepackage{hyperref}
\usepackage{url}

\usepackage{hyperref}
\usepackage{url}
\usepackage{graphicx} 
\usepackage{booktabs}
\usepackage{colortbl}
\usepackage{amssymb}
\usepackage{bbding}
\usepackage{pifont}
\usepackage{amsthm}
\usepackage{chngcntr}
\counterwithout{table}{section}
\usepackage{amsmath}

\newtheorem{theorem}{Theorem}

\newtheorem{proposition}[theorem]{Proposition}

\newtheorem{property}{Property}

\theoremstyle{remark}
\newtheorem{remark}{Remark}

\theoremstyle{definition}

\title{Latent Space Disentanglement in Diffusion Transformers Enables Precise Zero-shot Semantic Editing}

\author{Zitao Shuai, Chenwei Wu\thanks{co-first author}, Zhengxu Tang\thanks{co-first author}, Bowen Song \& Liyue Shen \thanks{coresponding author} \\
University of Michigan\\
\texttt{\{liyues\}@umich.edu}
}

\def\month{MM}  % Insert correct month for camera-ready version
\def\year{YYYY} % Insert correct year for camera-ready version
\def\openreview{\url{https://openreview.net/forum?id=XXXX}} % Insert correct link to OpenReview for camera-ready version

\begin{document}

\maketitle

\begin{abstract}
Diffusion Transformers (DiTs) have recently achieved remarkable success in text-guided image generation. In image editing, DiTs project text and image inputs to a joint latent space, from which they decode and synthesize new images. However, it remains largely unexplored how multimodal information collectively forms this joint space and how they guide the semantics of the synthesized images. In this paper, we investigate the latent space of DiT models and uncover two key properties: First, DiT's latent space is inherently semantically disentangled, where different semantic attributes can be controlled by specific editing directions. Second, consistent semantic editing requires utilizing the entire joint latent space, as neither encoded image nor text alone contains enough semantic information. We show that these editing directions can be obtained directly from text prompts, enabling precise semantic control without additional training or mask annotations. Based on these insights,  we propose a simple yet effective Encode-Identify-Manipulate (EIM) framework for zero-shot fine-grained image editing. Specifically, we first encode both the given source image and the text prompt that describes the image, to obtain the joint latent embedding. Then, using our proposed Hessian Score Distillation Sampling (HSDS) method, we identify editing directions that control specific target attributes while preserving other image features. These directions are guided by text prompts and used to manipulate the latent embeddings.
Moreover, we propose a new metric to quantify the disentanglement degree of the latent space of diffusion models. Extensive experiment results on our new curated benchmark dataset and analysis demonstrate DiT's disentanglement properties and effectiveness of the EIM framework. 
\footnote{Our annotated benchmark dataset is publicly available at https://anonymous.com/anonymous/EIM-Benchmark.}
% to facilitate the reproducible research in this domain.
\end{abstract}

\section{Introduction}
\begin{figure*}[t!] 
    \centering
    
    %\vspace{-15pt}
    \includegraphics[width=\textwidth]{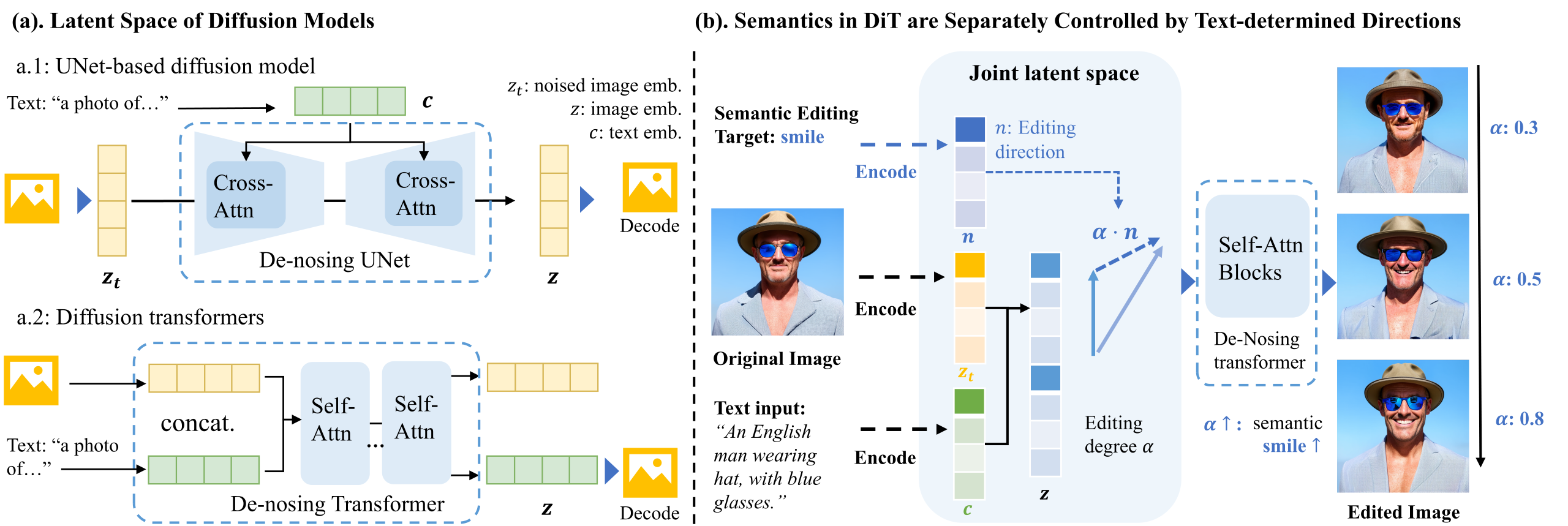} % Adjust the path and width as needed
    %\vspace{-15pt}
    \caption{(a) UNet-based models align text embeddings with image embeddings via cross-attention layers. In contrast, DiT creates a joint latent space by combining the text embedding and image embedding, then feeds them into the denoising block. (b) DiT's has semantic disentangled latent space, where intensities of image semantics in generated images are controlled by separate directions, which can be easily identified.
    }
    \vspace{-15pt}
    \label{figure1}
\end{figure*}

Diffusion models have achieved remarkable success in text-guided generation tasks, producing diverse, high-fidelity images and videos based on text prompts~\citep{ma2024latte, esser2024scaling}. These diffusion-based models are typically built on the UNet-based latent diffusion framework~\citep {rombach2022high}, where input images are projected into latent embeddings and integrated with text embeddings through cross-attention layers within the UNet, as shown in Fig.~\ref{figure1}(a.1). Recently, Diffusion Transformers (DiT) (Fig.~\ref{figure1}(a.2)) introduced a new architecture that combines input image and text embeddings into a joint latent space and processes them through stacked self-attention layers. While empirical results suggest that this new structure enhances T2I controllability~\citep{esser2024scaling}, it is unclear how multimodal information collectively forms this joint latent space and how these representations guide the semantics of the synthesized images.

% Therefore, existing works rely on the loss-driven methods or mask annotations approach to achieve controllable text-guided image generation, either by seeking semantic latent spaces where directions that control image semantics can be identified by loss-driven methods~\citep{kim2022diffusionclip,kwon2022diffusion,avrahami2022blended,mokady2023null,kawar2023imagic}, or utilizing intermediate features that connect image semantics with text inputs~\citep{hertz2022prompt,brooks2023instructpix2pix}. However, these methods often require additional annotations (e.g., sub-image masks) or extensive optimization and model tuning, making it challenging to identify and interpret the semantic latent space and less generalizable in a variety of tasks. Instead, DiT brings the new structure of self-attention layers to project image and text inputs into a joint latent space, which potentially allows image semantics to be directly linked to the text prompt. With this new design of attention mechanism as well as the remarkable results of T2I generation of DiT models, we are curious to ask: \textit{Is the joint latent space of DiT semantically disentangled?} TOO LONG!!!
A key to achieving controllability is the property of \textit{semantic disentanglement}, where semantic attributes are encoded into distinct subspaces that can be independently controlled along specific editing directions. In other words, we can precisely manipulate a semantic attribute without affecting the others. 
This property has not been clearly observed in the latent space constructed in the UNet-based diffusion models~\citep {lu2024hierarchical,kwon2022diffusion}.
Consequently, existing approaches alternatively rely on either loss-driven methods to identify semantic control directions~\citep{kim2022diffusionclip,kwon2022diffusion,avrahami2022blended,mokady2023null,kawar2023imagic} or attention maps to connect image semantics with text inputs~\citep{hertz2022prompt,brooks2023instructpix2pix}. 
Note that these methods all require additional annotations (e.g., sub-image masks) or extensive optimization to achieve precise semantic control in image editing task, limiting their interpretability and generalizability in broader applications. In contrast, DiT's self-attention architecture projects image and text inputs into a joint latent space, potentially enabling direct link between image semantics and text prompts. Given this novel attention mechanism and DiT's remarkable text-to-image generation performance~\citep{esser2024scaling}, we are motivated to delve into the fundamental research question: \textit{Is the joint latent space of DiT semantically disentangled?}

% We start by modifying image semantics in the joint latent space, with directions given by simply encoding a text prompt or a reference image, and have found two disentanglement properties: (1) The semantic attributes in the image can be edited through the specified editing directions in a controllable manner respectively in the latent space; (2) Modifying a semantic along its controlling direction in the latent space, does not impact other semantics. These properties allow for modifying specific semantics in fine granularity without affecting the others, with editing directions that can be easily identified.
% As shown in Fig.~\ref{figure1}(b), given an image input and its text description, we can identify the controlling semantic direction for a specific image attribute (e.g., 'smile') by encoding the corresponding text prompt for the target attribute (e.g., a single world 'smile'). This enables precise control over the intensity of the image semantic attribute by manipulating the joint latent embedding along this direction.

By modifying image semantics in the joint latent space using directions from encoded text prompts or reference images, we discover two disentanglement properties: (1) Semantic attributes can be edited through specific directions in a controllable manner; (2) Modifying semantics along one controlling direction does not affect other attributes. 
These properties enable precise and fine-grained modification of specific semantics while preserving others, using easily identifiable editing directions.
As shown in Fig.~\ref{figure1}(b), given an input image and its text description, we can identify the controlling direction for a specific attribute (e.g., `smile') by encoding its corresponding text prompt (e.g., the word `smile'). This enables precise control over the semantic intensity by manipulating the joint latent embedding along this direction.
 %More rigorously, in the DiT models, we define the joint latent space to be constructed directly by both the text embedding and noised image embedding, and thus the editing direction can be easily identified. %Unlike previous works, the joint latent space and the directions that control image semantics can be easily identified. As it is directly constructed by the text embedding and noised image embedding, which serve as inputs to the denoising transformer.
 
 These properties enable DiT to serve as a prior knowledge for precise semantic editing. Based on this, we propose a simple yet effective \textbf{E}ncode-\textbf{I}dentify-\textbf{M}anipulate (EIM) framework for zero-shot image editing with precise and fine-grained semantic control, while not requiring additional training or mask annotations. Given an input image and target semantic, we first extract the image's text description using a multi-modal LLM, then encode both image and text into a joint latent embedding. To precisely edit the target semantic, we introduce a Hessian-Score-Distillation-Sampling (HSDS) method that identifies disentangled editing directions guided by the target semantic's text prompt, as well as the desired editing degree. 
 The joint latent embedding is then linearly manipulated along this direction.
To evaluate our approach, we propose a \textbf{S}emantic \textbf{D}isentanglement m\textbf{E}tric (SDE) for measuring latent space disentanglement in text-to-image diffusion models. We also introduce ZOPIE (\textbf{Z}ero-shot \textbf{O}pen-source \textbf{P}recise \textbf{I}mage \textbf{E}diting), a new benchmark for comprehensive evaluation. Both automatic and human evaluations demonstrate our framework's effectiveness in precise image editing tasks.
% These properties allow for leveraging DiT as prior knowledge for precise image semantic editing. For this purpose, we propose a novel and simple \textbf{E}ncode-\textbf{I}dentify-\textbf{M}anipulate (EIM) framework that enables fine control over image semantic intensity, without any additional training or mask annotation. Given an image and a target semantic to be edited, we first extract the text description of the input image via a multi-modal LLM, and encode both image and text to form the joint latent embedding. To precisely edit the given target semantic, we propose a Hessian-Score-Distillation-Sampling (HSDS) method to identify the disentangled editing direction that only modifies the target semantic. This process is guided by the direction given by encoding the text prompt of the target semantic, and the desired editing degree. Finally, we linear manipulate the joint latent embedding with the identified editing direction. We additionally propose a \textbf{S}emantic \textbf{D}isentanglement m\textbf{E}tric (SDE) to measure the disentanglement property of the latent spaces of T2I diffusion models. In the absence of comprehensive precise image-editing evaluation datasets, we construct a new benchmark named as ZOPIE, a novel \textbf{Z}ero-shot \textbf{O}pen-source \textbf{P}recise \textbf{I}mage \textbf{E}diting benchmark. Both automatic and human evaluations demonstrate the effectiveness of our framework on precise editing tasks.

Our contributions are summarized as follows:
\begin{itemize}
\vspace{-5pt}
    \item We systematically studied and modeled the latent embedding space of DiT models, and propose a new metric to quantify its \textit{disentanglement}. Our study for the first time comprehensively uncover this important \textit{disentanglement} property of the DiT model, which set the foundation for controllable image editing with DiT models.

    \item We propose a simple yet effective Extract-Indentify-Manipulation (EIM) method via Hessian Score Distillation Sampling method for zero-shot image editing with precise and fine-grained semantic control, without requiring any additional training or mask annotations. 
    % EIS modifies image semantics in DiT's latent space, via manipulating latent embeddings along directions that can be easily identified.

    \item We introduce a novel \textbf{Z}ero-shot \textbf{O}pen-source \textbf{P}recise \textbf{I}mage \textbf{E}diting (ZOPIE) benchmark, with both automatic and human annotations to assess the effectiveness of precise image editing tasks, as well as the disentanglement property of generative models.
\end{itemize}

\begin{figure*}[t!] 
    \centering
    
    \includegraphics[width=1\textwidth]{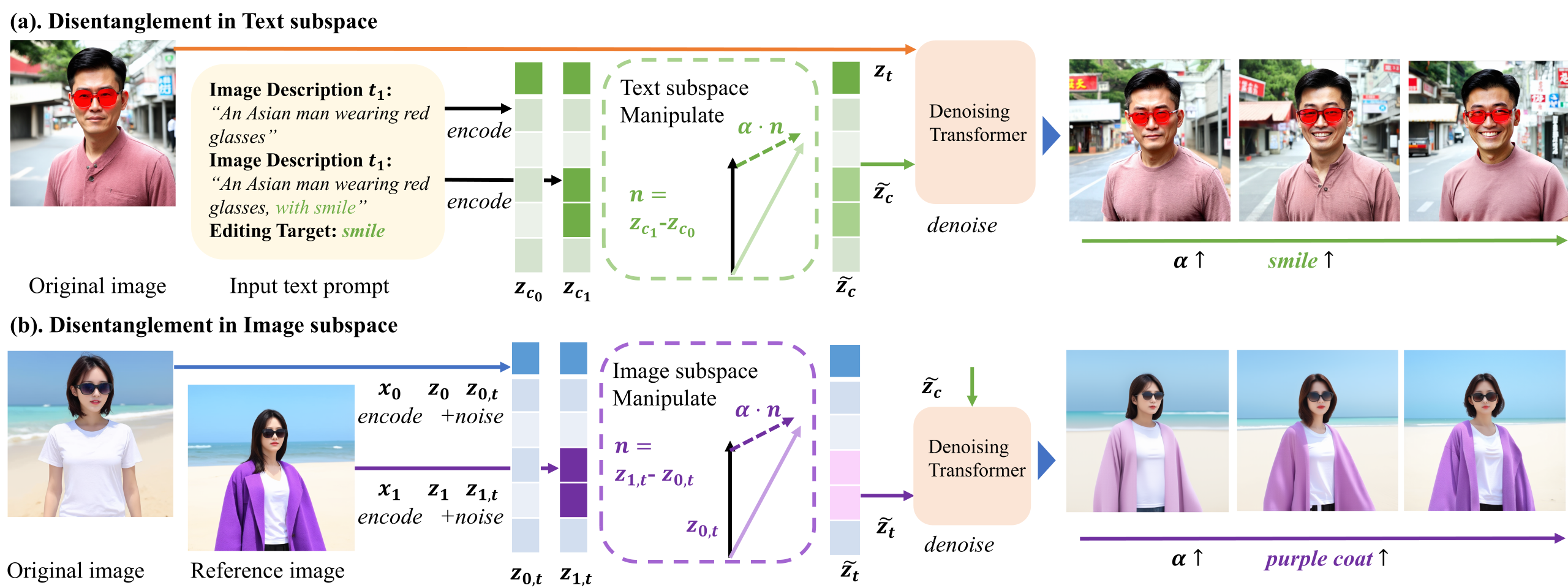} % Adjust the path and width as needed
    %\vspace{-10pt}
    \caption{The disentanglement properties of the joint latent space of DiT enable: (a) Given a target semantic to be edited, we can identify a direction that allows fine-grained control of the intensity of this semantic; (b) With a reference image that only differs in the target semantic, we can obtain such direction in the image embedding space and achieve similar precise control.
    }

\label{figure2}
\vspace{-10pt}
\end{figure*}

\section{Exploration on Latent Space of Diffusion Transformer}
\label{exploration}
%In this section, we will first provide preliminaries of diffusion transformers especially for the task of text-guided image generation and image editing. Then we will present our key findings of semantic disentanglement properties of the diffusion transformer.%, as well as our findings why this semantic latent space of DiT models is disentangled. 

\subsection{Preliminary}
\label{Preliminary}

\label{latent space}
\textbf{Diffusion Transformer.} 
Text-to-Image (T2I) diffusion transformer utilizes~\citep{peebles2023scalable} a transformer model as the denoising backbone.
In specific, T2I diffusion transformer employs the encoder of a variational auto-encoder (VAE) model to encode the input image to latent embeddings, a VAE decoder to transform image embeddings back to the pixel space, and a set of text encoders to encode the input text prompt to text embeddings. Given an image and a text prompt, the text encoder will project the prompt into text embeddings $ z_c \in \mathbb{R}^{l\times d} $, where $l$ is the length of tokens in the text prompt, and $d$ is the hidden dimension. By averaging $z_c$ across the token dimension, we can obtain a pooled text embedding $ \bar{z_c} \in \mathbb{R}^{d}$ which indicates coarse-grained global semantics. Meanwhile, the image will be encoded into image embeddings $ z \in \mathbb{R}^{v\times d} $, where $v$ is the length of the patchified image tokens, and $d$ denotes the hidden dimension, which is the same as text latent embedding. At forward time-step $t$, $z$ will be added with noise $\epsilon_t$, forming noised image embeddings $z_t$. Before the de-noising loops, $z_t$ and $z_c$ are combined into a joint latent embedding and input into the de-noising transformer. Here, the image and text latent embedding dimension are chosen to be the same in order to project both modalities into the joint latent space. They are concatenated to obtain $z\in \mathbb{R}^{(v+l)\times d}$, creating a joint latent space $\mathcal{Z}$. Therefore, this joint latent space consists of two subspaces: the image latent space \( \mathcal{Z}_t \), formed by \( z_t \), and the text latent space \( \mathcal{Z}_c \), formed by \( z_c \). After de-noising loops, the model is expected to generate the latent embedding for the edited image, with the VAE decoder will decode the denoised image embedding to pixel space to get the edited image.

%%%%%%%%%%%%%%%%%%%%%%%%%%%%%%%%%%%%%%%%%%%%%%%

%%%%%%%%%%%%%%%%%%%%%%%%%%%%%%%%%%%%%%%%%%%%%%%

\label{sec:image editing}
\textbf{Image Editing}. The investigation and modeling of the disentanglement properties in DiT's latent space are mainly based on the image editing task. 
The goal of image editing is to modify certain semantic contents in the given source image $x$, while keeping other semantic attributes unchanged. 
In this paper, we primarily focus on text-guided image editing, as we often edit images based on text instructions in practical applications. Following~\cite{wu2023uncovering}, we model the image editing as the two following processes in the latent space. 
(1) Image forward process: given image embedding $z$ and timestep $t$, $z$ is added with noise $\epsilon_t$ to get noised image embedding $ z_t$. 
(2) Image reverse process: we input text prompts to text encoders to gain text embeddings, and pass them together with $z_t$ into the de-noising transformer to get a de-noised image latent $\tilde{z}$. Then we decode the $\tilde{z}$ to obtain the edited target image $\tilde{x}$. In this paper, to demonstrate the significance of the disentanglement property for controllable image generation, we apply frozen diffusion models to precisely edit targeted semantics. To formalize the goal of this paper: given a semantic $s$ and a source image $x$, \textit{we aim to change the intensity $\alpha$ of a given semantic $s$ reflected in the edited target image $\tilde{x}$, while avoiding incorrectly changing other semantics.} The key differences to the most related works are summarized in Table~\ref{table1}.

\subsection{Disentangled Semantic Representation Space of Diffusion Transformer}
\label{disentangle space sec}
We present our key findings on DiT's latent space, highlighting its unique disentanglement properties compared to related works (Table~\ref{table1}). We illustrate these properties through specific editing examples and our experiments demonstrate their consistency across various cases.

Suppose we are given a source image, a target semantic to be edited, and a text prompt $t_0$ that describes the image. The prompt $t_0$ will be encoder to text embedding \( z_{c_0} \), and the source image will be transformed to image embedding $z_t$, both of them are combined to form a joint latent embedding. This joint embedding serves as the input to the de-noising transformer and is straightforward to obtain and manipulate. In the following part, we will demonstrate the semantic disentanglement properties of this joint latent space, as well as how they can be leveraged for editing intensities of specific image semantics.

\begin{figure*}[t!] 
    
    \centering
    \includegraphics[width=1\textwidth]{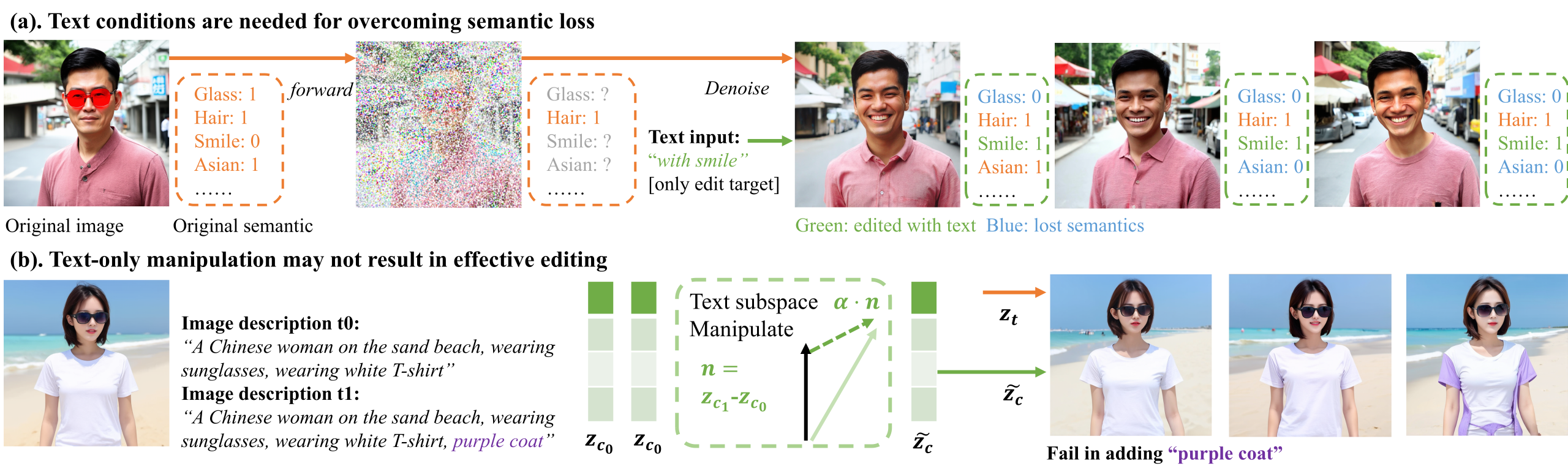} % Adjust the path and width as needed
    %\vspace{-10pt}
    \caption{To effectively modify semantics, we must manipulate the entire joint latent space. This is due to two observed challenges: (a) \textit{semantic loss}, where certain semantics are randomly recovered in the denoising process if they are not conditioned; (b) manipulating only text embeddings is insufficient for modifying certain semantics effectively.
    }
    \vspace{-15pt}
    \label{figure3}
\end{figure*}

\label{text disentangle}
\textbf{Image semantics can be manipulated separately by image or text in the joint latent space.} 
For a specific semantic content, we first would like to find out the controllable editing direction $n$ in $\mathcal{Z}_c$, which can precisely edit the intensity of this semantic reflected in the image, while leaving the others unchanged. Here, $n$ can be determined explicitly by encoding related text prompts. Specifically, given text prompt $t_1$ that describes the demanded image, and text prompt $t_0$ that describes the original image, we first encode them to obtain text embedding $z_{c_1}$ and $z_{c_0}$ respectively. We then subtract them to obtain $n$. %: $n = z_{c_1} - z_{c_0}$. Here, $z_{c_0}$ is encoded from the text description of the source image, and $z_{c_1}$ is from that of the desired image.
With such direction $n$, given an editing degree \( \alpha \), we suppose to manipulate \( z_{c_0} \) %encoded from the text description of the source image 
by applying \( \tilde{z_c} = z_{c_0} + \alpha \cdot n \). As shown in Fig.~\ref{figure2}, this approach has successfully modified the intensity of the semantic ``smile'' with fine granularities while avoiding affecting other semantics.

\label{image disentangle}
We have observed similar phenomenons in the image latent space. Similar to the text side, we start from seeking the editing direction $n$. As shown in Fig.~\ref{figure2}, we carefully select two images $x_1, x_0$ that are mainly different in the semantic attribute "coat". We encode both of them to obtain image embedding $z_{1}$ and $z_{0}$, then we add noise $\epsilon_t$ to them to obtain $z_{1,t}$ and $z_{0,t}$. We found that the distance between $z_{1,t}$ and $z_{0,t}$ can be utilized as the editing direction of the semantic "coat". Therefore, we suppose the editing direction $n$ can be given by $z_{1,t} - z_{0,t}$. We linearly manipulate the image embeddings following $\tilde{z_{t}}= \alpha \cdot n +  z_{0,t}$, where the image embedding $z_{0,t}$ of the image without "coat" shifted toward desired $z_{1,t}$ with different ratio $\alpha$. 
As $\alpha$ increases, the "coat" semantic gets more significant. 

\begin{figure}
    \centering
    \includegraphics[width=1\linewidth]{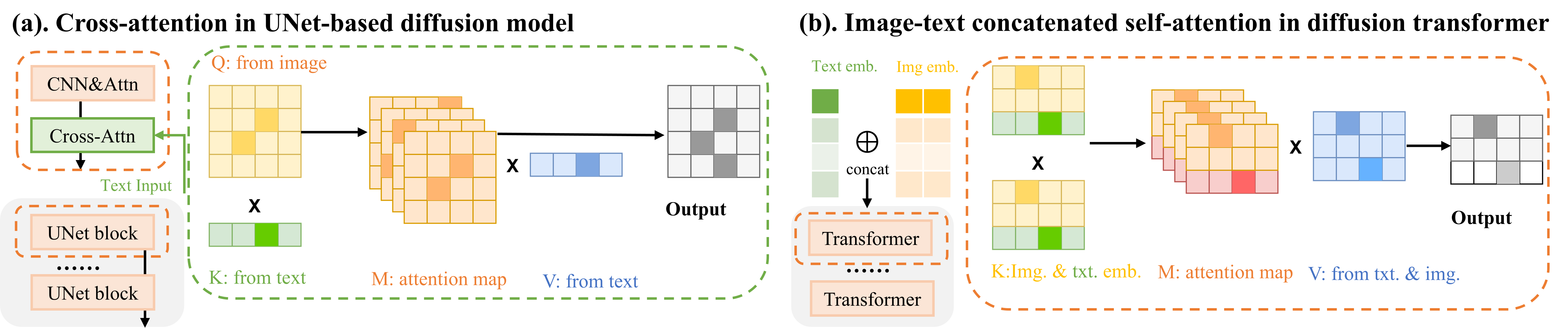}
    \caption{(a) Image-text cross-attention mechanism in UNet-based diffusion models. (b) Self-attention mechanism in the diffusion transformer, where image and text embeddings are concatenated and jointly flowed into the attention block. In the self-attention mechanism, the attention map of a specific semantic contains less category information of other semantics, as shown in Sec.~\ref{sec: probing}. }
    \vspace{-15pt}
    \label{fig:attention}
\end{figure}

The observations discussed above indicate that image semantics might be encoded into different subspaces, where their intensities are controlled by separate directions. Meanwhile, these directions \textit{can be straightforwardly identified}, either by encoding related text prompts or projecting reference images, if a reference image exists. This disentanglement property may come from the nature that attention maps of semantics in the diffusion transformer are less entangled, as shown in Fig.~\ref{fig:attention}. These findings may unlock the potential of adjusting the intensity of a given semantic, through linearly manipulating latent embeddings near the boundary defined by the editing direction. Detailed theoretical analysis can be found in Appendix~\ref{sec:feasibility}.

%\textbf{$\mathcal{Z}$ satisfies both properties and thus is a disentangled semantic representation space.} We have verified the decomposability property of both $\mathcal{C}$ and $\mathcal{Z}_t$, thus their joint space $\mathcal{Z}$ is also decomposable. The effectiveness of $\mathcal{Z}$ has been empirically verified in experiments sections. Therefore, $\mathcal{Z}$ is disentangled, allowing image semantics to be encoded into different subspaces, where their intensities are controlled by separate directions. This enables linearly adjusting the intensity of a given semantic, through linearly manipulating latent embeddings near the boundary defined by the editing direction. Detailed theoretical analysis can be found in Appendix~\ref{sec:feasibility}. 

\textbf{Achieving effective semantic editing across different targets requires utilizing the entire joint latent space.} While semantic attributes are encoded in the joint latent space through image and text latent embedding seperately as described in the last section, neither modifications on $\mathcal{Z}_c$ nor $\mathcal{Z}_t$ alone can effectively modify all desired semantics in a precise control manner. Firstly, during the forward process, $\mathcal{Z}_t$ may suffer from the `Semantic Loss Phenomenon'~\cite{yue2024exploring}, losing some of the semantics $\mathcal{Z}_0$ contained. 
As demonstrated in Fig.~\ref{figure3}, in the forward process, some attributes become noised; they will be assigned random values if not conditioned by text embeddings during the reverse process. For example, in Fig.~\ref{figure3}, when the semantics `ethnicity' and `glasses' are left unconditioned on the text side, the reconstructed image samples exhibit random race and glasses attributes. In contrast, when conditioned by text embeddings, the attributes are correctly recovered, as shown in Fig.~\ref{figure2}. This observation indicates that some semantics cannot be effectively conditioned if text embeddings are not well leveraged, suggesting that $\mathcal{Z}_t$ alone is insufficient for effective control.

Conversely, manipulating only $\mathcal{Z}_c$ fails to edit certain semantics. Fig.~\ref{figure3} demonstrates this limitation when attempting to modify a 'no coat' image to include a 'purple coat', where we can see by only editing the text prompt while not manipulate the image embedding, the model fails to add a correct `purple coat' as desried.
As shown in Fig.~\ref{figure2}(b), the edit succeeds only when we manipulate both the image and text latent spaces by reweighting $z_t$ of the source image along with reference images containing purple coats. Here we assume the access to the reference image for the purpose of analysis. However, the reference image, which is actually the editing target image, does not exist in practical image editing tasks. In the following Method section, we will discuss how to manipulate $z_t$ without reference images.

\section{Method}
%In this section, we will first propose our \textbf{E}ncode-\textbf{I}dentify-\textbf{M}anipulation (EIM) framework for precise image editing, which is motivated by the disentanglement properties of the diffusion transformer. Then we will introduce our \textbf{S}emantic \textbf{D}isentanglement m\textbf{E}tric (SDE) metric for measuring the disentanglement degree.

\subsection{Encode-Identify-Manipulate Framework}
\label{sec  EIS }
% In this section, we will first propose our \textbf{E}ncode-\textbf{I}dentify-\textbf{M}anipulate (EIM) framework for precise image editing, which is motivated by our found disentanglement properties of the diffusion transformer. The whole framework can be seen in Fig.~\ref{fig:method}.

\begin{figure*}[t!] 
    \centering
    \includegraphics[width=1\textwidth]{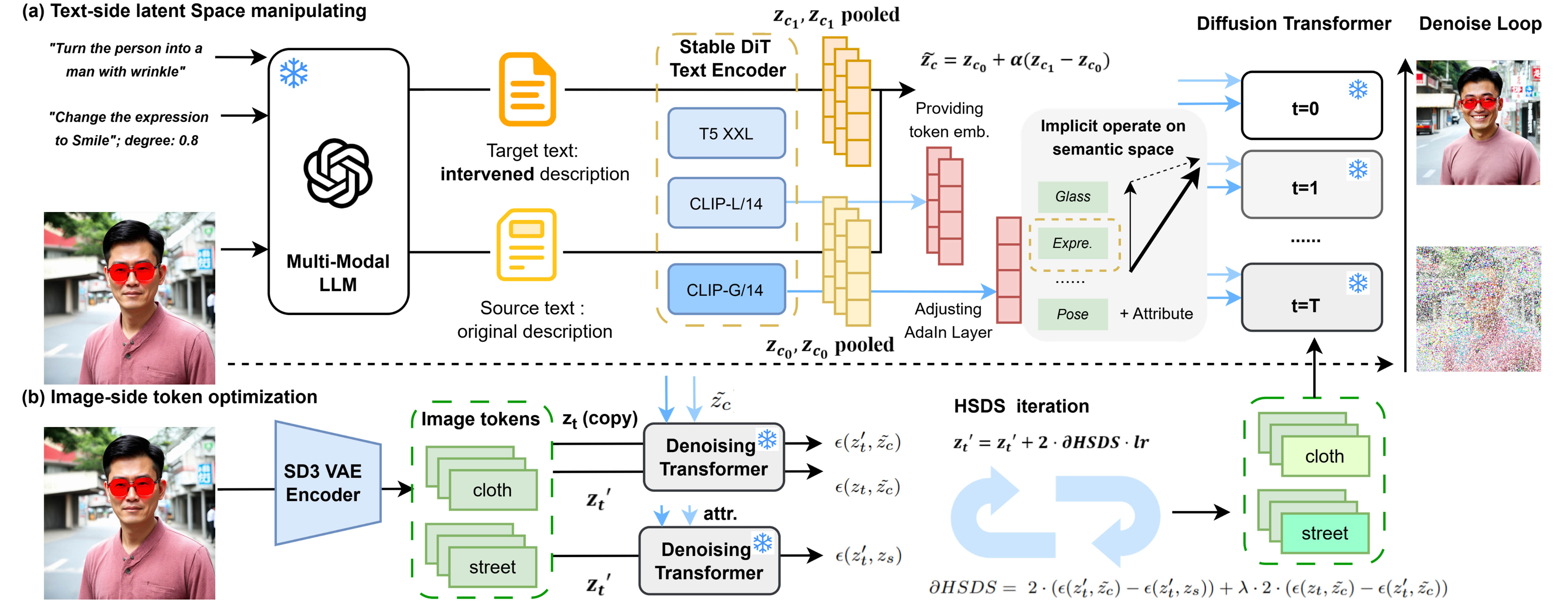} % Adjust the path and width as needed
    %\vspace{-10pt}
    \caption{(a). \textbf{Encode}: Given a source image, we first utilize a multi-modal LLM to get a source prompt that describes semantics of the image, and a target prompt for the desired image. We encode the source prompt and the given image to a joint latent embedding. (b). \textbf{Identify}: We obtain the editing direction in text subspace by subtracting embeddings encoded from the source prompt and target prompt. We propose a Hessian-Score-Distillation-Sampling method to identify the editing direction in image subspace. (c). \textbf{Manipulate.} Finally, we linearly combine the joint latent embedding with on the identified direction.
    }
    \vspace{-10pt}
    \label{fig:method}
\end{figure*}

% Suppose we are given a source image $x$ and the target semantic $s$ (e.g., 'smile') to be edited. In the precise image semantic editing task, we aim to modify the intensity of $s$ reflected in the edited image. Specifically, considering the image-text joint self-attention mechanism in DiT's denoising transformer, we also include the text description of source image, denoted as $t_0$ to help form the joint latent embedding. We also need a $t_1$ that describes the desired image for identifying the editing direction. In the case where we do not have $t_0$ and $t_1$, we will use a multi-modal LLM to generate both of them
Motivated by the disentanglement properties of diffusion transformers, we propose our \textbf{E}ncode-\textbf{I}dentify-\textbf{M}anipulate (EIM) framework for precise image editing (Fig.~\ref{fig:method}).
Given a source image $x$ and target semantic $s$ (e.g., 'smile'), our goal is to precisely control the intensity of $s$ in the edited image. 

The framework requires both the source image's text description $t_0$ and the desired image's description $t_1$ to form the joint latent embedding, leveraging DiT's image-text self-attention mechanism. When these descriptions are unavailable, we generate them using a multi-modal LLM.
\footnote{In our method, we instruct the multi-modal LLM to generate text descriptions for both the original and desired images, ensuring that they differ only in the tokens corresponding to the semantics to be edited. Consequently, in the encoded text embeddings, only parts related to the specified semantics are altered.}.

\textbf{Encode.} %To accurately only modify target semantics, we first project the source image $x$ and its text description $t_0$ to form a joint latent embedding. 
As demonstrated in Sec~\ref{disentangle space sec}, we have shown that when the task-irrelevant semantics are not conditioned during the denoising process, they will be randomly changed due to the semantic loss phenomenon. Thus, we need to inject the text embedding $z_{c_0}$ of text prompt $t_0$, which describes the source image, as the condition to help correctly maintain these semantics. Meanwhile, the given image $x$ is encoded into image embedding $z$, and is added with a randomly sampled noise $\epsilon_t$ at timestep $t$, to obtain the noised image embedding $z_t$. And we also encode the text description (e.g., 'purple coat') of our editing target to a text embedding $z_s$.

%That's because semantics would get lost in the forward process, and are randomly reconstructed in the reverse process if they are not conditioned with text embeddings, as shown in Sec.~\ref{disentangle space sec}. 

%To guide the recovery of these semantics, given the input image \(x\), we first obtain its text description \(t_0\) using a multi-modal LLM~\cite{achiam2023gpt} , as shown in Fig.6. 
%Based on the generated image captioning for the source image, we edit the words/expression... to be corresponding to the target text...reflect the desired semantic change in the target image....
%We denote this edited text as $t_1$, which is also the prompt for generating the target image.
%Similarly, we generate a text prompt \(t_1\) for the desired image\footnote{In our method, we instruct the multi-modal LLM to generate text descriptions for both the original and desired images, ensuring that they differ only in the tokens corresponding to the semantics to be edited. Consequently, in the encoded text embeddings, only parts related to the specified semantics are altered.}, through instructing the LLM to only edit words related to the target semantic in \(t_0\). 

\textbf{Identify.} In order to manipulate the joint latent embedding for editing target image semantics, we aim to identify an editing direction that controls this semantic. Specifically, since $\mathcal{Z}$ consists of subspaces $\mathcal{Z}_t$ and $\mathcal{Z}_c$, we can separately obtain the editing direction for each subspace. For text latent space, the editing directions are explicitly associated with text prompts. Therefore, given the ratio $\alpha$ for modifying the semantic $s$, we can firstly encode the text $t_1$ which describes the desired image, to obtain text embedding $z_{c_1}$. Then we construct the editing direction $n$ here by $n_c=\alpha (z_{c_1} - z_{c_0})$. Therefore, we can operate the text embedding with: $\tilde{z_c}=z_{c_0}+n_{c}$.

In practice, for the noised image embeddings $z_t$, assuming that there is no reference image, we lack an explicit editing direction as in the text side. To address this problem, we obtain this direction based on a score distillation sampling~\citep{hertz2023delta,poole2022dreamfusion} method, which computes an update direction $n_{z_t}$ for latent image embedding $z_t$. During the de-noising process, we have observed that de-noising with
$z_{s}$ that encoded from the text prompt of the desired semantic (e.g., 'purple coat'), leads to higher intensity of this semantic in the generated image. Thus, we assume the difference of the two predicted noise conditioned on $z_{s}$ and $z_{c_0}$, can offer a rough direction for modifying the intensity of this semantic. Therefore, it is straightforward to %utilize $\partial \|\epsilon_{\text{attr}}(z_t) - \epsilon_t\|_2$ as the update direction, or 
employ delta distillation sampling~\citep{hertz2023delta} to push image embedding $z_t$ to $z_t'$ along the rough editing direction, which follows
$\partial (\|\epsilon(z_t',z_s) - \epsilon_t\|_2 - \|\epsilon(z_t,z_{c_0}) - \epsilon_t\|_2)$ as the update direction. Here $\epsilon_t$ is a random noise sampled at time-step\footnote{Different from SDS that focus on optimizing image embedding $z$, we aim to utilize SDS-based method to seek an editing direction for noised image embedding $z_t$ at timestep $t$, thus we hold a fixed time-step $t$ when utilizing SDS-based method.} $t$.

However, this might incorrectly affect the semantics controlled by text embeddings. In contrast, we should encourage the image embedding moves in a precise direction, to avoid affecting semantics irrelevant to the editing task. Motivated by this insight, we propose a Hessian Score Distillation Sampling (HSDS) method, which aims to minimize the distance between $\epsilon(z_t',\tilde{z_c})-\epsilon(z_t,z_{c_0})$ and $\epsilon(z_t',z_s)-\epsilon(z_t,z_{c_0})$. This aligns $\epsilon(z_t',\tilde{z_c})$ and $\epsilon(z_t', z_s)$ alongside the direction that can increase the intensity of the target semantic, while mitigating unintended changes on other semantics. Meanwhile, we add a constrained term which aims to reduce the distance between $\epsilon(z_t',\tilde{z_c})$ and noise $\epsilon(z_t, \tilde{z_c})$ predicted using $z_t$ and $\tilde{c}$ as input. Finally, the image gradient of the Hessian Score-Distillation-Sampling should be:
\begin{equation}
\footnotesize
\label{csds}
    \partial HSDS=\ 2\cdot(\epsilon(z_t', \tilde{z_c})-\epsilon(z_t',z_s))+\lambda\cdot 2\cdot(\epsilon(z_t,\tilde{z_c})-\epsilon(z_t', \tilde{z_c}))
\end{equation}
, where $\lambda$ is the hyper-parameter that adjusts the degree to which the manipulated image embedding $z_t'$ should be close to the original $z_t$. Detailed analysis can be seen in appendix~\ref{sec theory HSDS}.
 
In score-distillation-sampling, we do not require the gradient of the diffusion transformer. Hence we can directly compute $\partial HSDS$, and update the image embedding with: $z_t'=z_t + n_{z_t}, n_{z_t}= \eta \cdot \partial HSDS$, where $\eta$ refers to the step size for the update loop. For implementation, we iteratively update $z_t$ to obtain a more accurate $n_{z_t}$. 

\textbf{Manipulate.} We manipulate the joint latent embedding through separate linear combinations in the text and image subspaces, with the identified editing direction $n_{c}$ and $n_{z_t}$. For the text side, we operate the text embedding with: $\tilde{z_c}=z_{c_0}+n_{c}$. Similarly, we operate $z_t$ with $\tilde{z_t}=z_t + n_{z_t}$.

\subsection{Semantic Disentanglement Metric}
\label{SDE}
Based on disentanglement properties demonstrated in Sec.~\ref{exploration}, we propose a metric that can be automatically computed to measure the degree of disentanglement of the latent space of text-to-image diffusion models. As we mentioned in Sec.~\ref{disentangle space sec}, a disentangled semantic representation space $S'$ should satisfy decomposability and effectiveness properties, where we can locally manipulate certain subspaces to effectively modify specific semantics without altering other semantics. Building on this insight, we propose to evaluate the disentanglement degree of a semantic representation space using an image-editing-based approach. Specifically,  we aim to examine the following properties\footnote{Detailed modeling and analysis of these properties can be seen in the appendix~\ref{sec formal disentanglement}.}: (1). \textbf{Effectiveness}: Can we effectively edit certain semantics in the latent space? (2). \textbf{Decomposable}: Manipulating latent embedding along editing direction of a specific semantic, would not affect the other semantics? 

Consider an image $x$ with a binary semantic $s\in [0,1]$ that will be lost in the forward process, which is labeled as $y_0=0$. We encode $x$ to latent space to get image embedding $z$, and forward $z$ to obtain $z_t$. We then denoise $z_t$ in the reverse process, and decode it to get the reconstructed image. To ensure effective editing, the reconstructed image \( \tilde{x} \), conditioned on \( y_1 = 1 \), should differ significantly from \( \hat{x} \), which is recovered under the condition \( y_0 = 0 \). This difference is measured by comparing their respective distances to the original \( x \). Simultaneously, to maintain the decomposability property, the distance between \( x \) and \( \tilde{x} \) should be small, ensuring that task-irrelevant semantics remain unchanged during the editing process.
\begin{remark}
    Given a diffusion model with a encode $f$ which turns a given image into the noised image embedding, and a decode function $h$ which turns noised image embedding to the recovered image, an image labeled with $y\in[0,1]$ of semantics $s$, and text embeddings $c, \tilde{c}$ encoded from text prompts correspond to $y$ and $1-y$. The \textbf{S}emantic \textbf{D}isentanglement m\textbf{E}tric (SDE) could be written as: 
    \begin{equation}
    \footnotesize
    \text{SDE}=\frac{||x-h(f(x,t), c, t)||_2}{||x-h(f(x,t), \tilde{c},t)||_2}+||x-h(f(x,t), \tilde{c}, t)||_2
    \end{equation}
    where $t$ is the forward time-step.
\end{remark}

\section{Experiment}
\subsection{Experiment Setting}
\textbf{Implementation Detail}. We include three state-of-the-art open-source transformer-based Text-to-Image (T2I) diffusion model for our experiments: Stable Diffusion V3~\cite{esser2024scaling}, Stable Diffusion V3.5\footnote{https://huggingface.co/stabilityai/stable-diffusion-3.5-medium}, and Flux\footnote{https://github.com/black-forest-labs/flux}. We follow their implementation in the Huggingface Diffusers platform~\cite{von-platen-etal-2022-diffusers}. During image editing tasks, we set the classifier-free-guidence (CFG) scale to 7.5 and the total sampling steps to 50. We forward the image to $75\%$ of the total timesteps, and then conduct the reverse process. For UNet-based T2I model, we include the widely utilized Stable Diffusion V2.1 from Huggingface Diffusers. We set the hyperparameters in the same way as those mentioned above. During our experiments, we keep diffusion models frozen. We utilize the GPT-4~\cite{achiam2023gpt} to extract text descriptions of given images. Details can be found in the appendix~\ref{sec mmllm} and~\ref{sec imple details}.

\textbf{Baselines.} Text-guided zero-shot precise image editing is a novel task; we adapt and compare to existing text-guided image editing methods DiffEdit~\cite{couairon2022diffedit}, Pix2Pix~\cite{brooks2023instructpix2pix}, and MasaCtrl~\cite{cao2023masactrl} as our baselines. To enable precise image editing, we apply similar text manipulation operations to the text embeddings that guide the reverse process in image editing.  We also include baseline methods that utilize the diffusion transformer but modify text embeddings only or manipulate image embeddings only.

\textbf{Qualitative Evaluation.} We conduct a comprehensive qualitative assessment of EIM using a diverse array of real images spanning multiple domains. Our evaluation employ simple text prompts to describe various editing categories, including but not limited to style, appearance, shape, texture, color, and lighting. These edits are applied to a wide range of objects, such as humans, animals, landscapes, vehicles, and surfaces. To demonstrate the necessity of image-side manipulation, we compare  EIM's performance with text-only embedding space manipulations. Table~\ref{tab:table2} shows a checklist of the qualitatively investigated image editing tasks along with illustrative examples. For our source images, we generate high-resolution samples using diffusion transformers. We then apply our method with three different editing degrees to showcase precise editing capabilities and the controllability of the editing degree. 

Additionally, we compare the precise editing results of our EIM method with those of the baseline models. We adapt the baselines to the same experimental pipeline for precise image editing. Like EIM, these methods are applied with three different editing degrees to demonstrate their capabilities. %The comparison highlights EIM's superior ability to achieve precise, precise edits and control the degree of modification.

\textbf{Quantitative Evaluation.} To address the lack of evaluation benchmarks for precise zero-shot image editing, we introduce a Zero-shot Open-source Precise Image Editing (ZOPIE) benchmark. ZOPIE includes both human subjective and automatic objective evaluations, focusing on image quality, controllability, image-text consistency, background preservation, and semantic disentanglement. The benchmark consists of 576 images across 8 editing types and 8 object categories, with annotations for source prompt, edit prompt, editing feature, object class, region of interest, and background mask. Details are provided in the appendix~\ref{sec imple details}.

For quantitative evaluation, we employed six metrics across four main aspects: 1) Background preservation: PSNR, LPIPS, and SSIM. 2) Text-image consistency: CLIPScore. 3) Editing effectiveness: Multimodal LLM-based Visual Question Answering score (MLLM-VQA score), which leverages multimodal language models to assess the smoothness and perceptibility of semantic intensity control.

The CLIPS score is calculated between the edit prompt and the edited images. For the MLLM-VQA score, pairs of images with consecutive editing strengths are fed into GPT-4 with the question: "Answer only in yes or no, does the second image reflect a gradual change of [edit feature] compared to the first image?" This process is repeated five times per sample, with the percentage of "yes" responses used to quantify the effectiveness of precise editing. Evaluation details can be seen in the appendix~\ref{sec imple details}.
\vspace{-8pt}

\textbf{Evaluation on semantic disentanglement degree.} We use the SDE metric proposed in Sec.~\ref{SDE} to assess the disentanglement degree of text-to-image diffusion models quantitatively. Specifically, we calculate the SDE for both the UNet-based diffusion model and the diffusion transformer using the CeleBA~\cite{liu2015deep} dataset—a widely used benchmark for evaluating generative capabilities in diffusion models~\cite{rombach2022high,zhang2024distributionally}. Each image in CeleBA is associated with an attribute vector, making it well-suited for measuring disentanglement with the SDE metric. We randomly sample 2,000 images and evaluated the models' disentanglement degree across age, gender, expression, hair, eyeglasses, and hat attributes.

\begin{table}[h] % The 't' means "top of the page"
\centering % This centers the table

\caption{Checklist of the qualitatively investigated editing task. Text tokens are more efficient in controlling semantics related to person and object than attribute and texture.}
\scalebox{0.71}{
\begin{tabular}{@{} l lll lll lll lll}
\toprule
\textbf{method} &\multicolumn{3}{c}{\textbf{Attribute}} &\multicolumn{3}{c}{\textbf{Texture}} &\multicolumn{3}{c}{textbf{Person}}  &\multicolumn{3}{c}{\textbf{Object}}\\
\cmidrule(l){2-4} \cmidrule(l){5-7} \cmidrule(l){8-10} \cmidrule(l){11-13} 
& Category & Example & &Category & Example & &Category & Example & &Category & Example & \\

\midrule
EIM & quantity& tiny$\to$huge& $\checkmark$& light & bright$\to$dark & $\checkmark$& smile & slight$\to$intense & $\checkmark$ & cover & none$\to$coat & $\checkmark$\\
EIM & shape & rect.$\to$spher.&$\checkmark$ & surface& uneven$\to$iron & $\checkmark$ & age & young$\to$old & $\checkmark$ & small item & pig$\to$cat &$\checkmark$\\
EIM & color & blue $\to$ green& $\checkmark$ & style & pencil$\to$painting & $\checkmark$ & wrinkle & severe$\to$slight & $\checkmark$ & large object & kangaroo$\to$bear & $\checkmark$\\

Text-only &quantity& tiny$\to$huge&  & light & bright$\to$dark &  & smile & slight$\to$intense & $\checkmark$ & cover & none$\to$coat &\\
Text-only & shape & rect.$\to$spher.& & surface& uneven$\to$iron & $\checkmark$& age & young$\to$old & $\checkmark$& small item & pig$\to$cat &$\checkmark$\\
Text-only& color & blue $\to$ green& $\checkmark$ & style&pencil$\to$painting & & wrinkle & severe$\to$slight & $\checkmark$& large object & kangaroo$\to$bear & $\checkmark$\\
% ... Your other rows here
\bottomrule
\end{tabular}
}
\label{tab:table2}
\end{table}

\subsection{Qualitative Results}
\label{sec:main result}

%\textbf{The joint latent space of DiT has achieved decomposability and effectiveness, and thus is disentangled.} We have systematically verified the decomposability of both image latent space and text latent space, on each type of semantic as shown in Table~\ref{tab:table2}. For each given semantic, we seek the image side and text side editing direction through our EIS method. As shown in Fig.\ref{}, all of the semantics can be effectively edited. when semantics are controlled by text subspaces, we can achieve precise and fine-granularity editing by only manipulating text embedding. Similarly, when controlled by image subspaces, precise editing can be achieved by adjusting only the image embedding. These results have empirically verified the decomposability and effectiveness of DiT's latent space.
\begin{figure*}[t!] 
    \centering
    
    \includegraphics[width=1\textwidth]{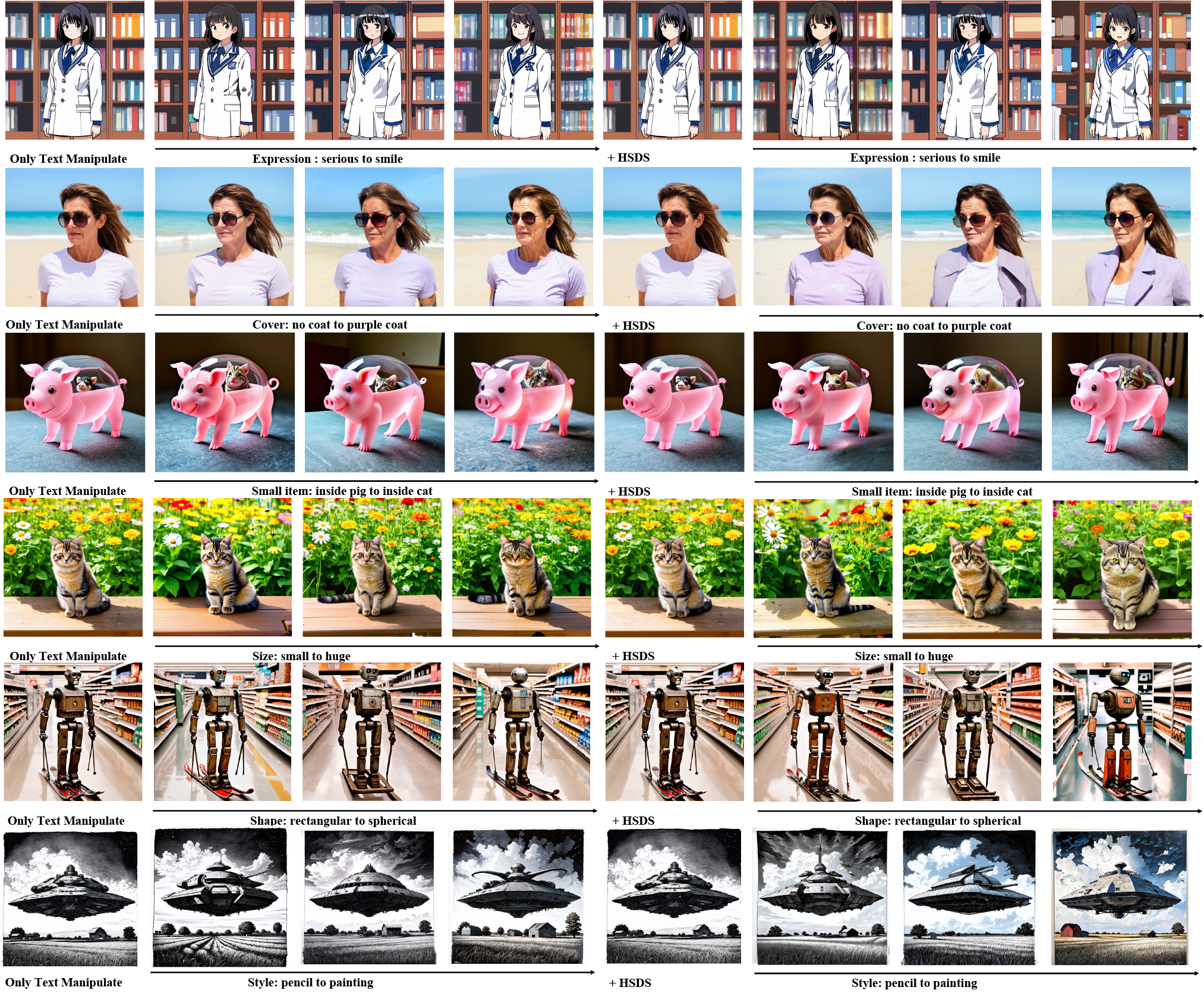} % Adjust the path and width as needed
    %\vspace{-10pt}
    \caption{Example precise image editing results of diverse semantics such as texture, shape, size, and expression. We utilize diffusion transformers to modify the intensity of considered attributes in fine granularity. The results on the left are from text-only manipulation, and on the right are from the complete EIM method.
    }
    \vspace{-10pt}
\label{ EIS}
\end{figure*}
\textbf{EIM achieves greater effectiveness in editing target semantics by operating on the entire semantic representation space.} Fig.\ref{ EIS} compares the images resulting from manipulating the entire joint latent embedding versus those from manipulating only the text embedding. While our EIM method can effectively edit target semantics by modifying the whole latent embedding, there are some failure cases in employing only text-side manipulation. For example, for editing tasks "enlarging a cat sitting in the garden" and "adding a purple coat to an American woman by the sea," the image subspace should also be manipulated for effective image semantic editing. More examples can be seen in the appendix~\ref{sec supplementary}.

\textbf{EIM outperforms baseline approaches in precise image editing and controls the intensity of target semantics in fine granularities.} As illustrated in Fig.~\ref{baseline}, the baseline methods frequently introduce unintended changes to irrelevant semantics, disrupting the overall image coherence. In contrast, our EIM method precisely modifies target semantics while keeping others unchanged, allowing for fine-grained adjustments to the intensity of the target semantics. More examples can be seen in the appendix~\ref{sec supplementary}.

\subsection{Quantitative Results}
\textbf{Our EIM method achieves better precise image editing results.} As shown in Table~\ref{quantitative table}, EIM demonstrates a significant advantage in background preservation during edits, with a strong PSNR of 20.8 and the lowest LPIPS score of 133.6. This showcases  EIM's ability to decompose and isolate specific semantics, ensuring that areas irrelevant to the editing target remain unaffected by the modifications. Moreover, the high CLIP Similarity Score of 34.3 achieved by EIM highlights its effectiveness in accurately editing target semantics. Finally, the significant advantage observed in the MLLM-VQA metric, where EIM scores 43.8, demonstrates its strength in fine-grained editing, reflecting the model's ability to produce nuanced and precise modifications.

\begin{table}[ht]

    \centering
    \caption{Quantitative Results}
    \resizebox{\textwidth}{!}{%
    \begin{tabular}{lcccccc}
        \toprule
        & \multicolumn{3}{c}{Background Preservation} & \multicolumn{1}{c}{CLIP Similarity Score} &\multicolumn{1}{c}{MLLM-VQA} \\
        \cmidrule(lr){2-4} \cmidrule(lr){5-5} \cmidrule(lr){6-6}
        Method & PSNR $\uparrow$ & LPIPS $\downarrow \times 10^3$ & SSIM $\uparrow \times 10^2$ & Whole $\uparrow \times 10^2$ & Whole $\uparrow \times 10^2$  \\
        \midrule
        P2P~\cite{brooks2023instructpix2pix} & \textbf{21.0} & \underline{139} & \underline{79.5} & 30.1 & 16.7 \\
        DiffEdit~\cite{couairon2022diffedit} & 19.1 & 205.4 & 69.9 & 30.7 & 14.6 \\
        MasaCtrl~\cite{cao2023masactrl} & 16.4 & 195.8 & 68.7 & 28.5 & 18.5 \\
        \hline
        Ours (Prompt Only) & 20.2 & 183.5 & 73.9 & \underline{31.2} & \underline{29.2} \\
        Ours (HSDS Only) & 18.9 & 190.2 & 71.5 & 30.9 & 27.4\\
        Ours & \underline{20.8} & \textbf{133.6} & \textbf{79.6} & \textbf{34.3} & \textbf{43.8} \\
        \bottomrule
    \end{tabular}%
    }
    \label{quantitative table}
\end{table}

\textbf{The diffusion transformer has better semantic disentanglement degree.} As indicated in Table~\ref{sde result}, the diffusion transformer consistently achieves lower SDE scores across various attributes such as age, gender, expression, hair, eyeglasses, and hat, compared to the UNet-based diffusion model. For example, the average SDE score of the stable diffusion 3 is significantly better compared to that of the UNet-based model. This suggests that the diffusion transformer has a more disentangled representation space, leading to more precise control during image editing.

\begin{table}[h]
\centering
\resizebox{\textwidth}{!}{
\begin{tabular}{l lllllll}
\toprule
Model & Backbone & SDE on Age$\downarrow$ & SDE on Gender$\downarrow$ & SDE on Expression$\downarrow$ & SDE on Hair$\downarrow$ & SDE on Eyeglasses$\downarrow$ & SDE on Hat$\downarrow$ \\
\midrule
SD2~\cite{rombach2022high} & UNet & 1.42 & 1.12 & 1.42 & 1.29 & 1.28 & 1.27 \\
SD3~\cite{esser2024scaling} & Transformer&  1.14 & 1.08 & 1.15 & 1.19 & \textbf{1.12} & 1.10 \\
SD3.5~\cite{} & Transformer& 1.11 & 1.12 & \textbf{1.10} & \textbf{1.18} & 1.13 & \textbf{1.07}\\
Flux~\cite{} & Transformer& \textbf{1.09} & \textbf{1.07} & 1.13 & 1.21 & 1.15 & 1.08 \\
\bottomrule
\end{tabular}
}
\caption{Semantic disentanglement metric of the diffusion transformer and the UNet-based diffusion model.}
\label{sde result}

\end{table}

\section{Analysis Study}
\label{sec: probing}
\subsection{Probing Analysis on Disentanglement Mechanism}
In this section, we will demonstrate the potential origin of the disentanglement properties of diffusion transformer's latent space, as well as why the representations of classical UNet-based diffusion models are always entangled.

We start by revisiting different text conditioning approaches of these two types of models. As shown in Fig~\ref{fig:attention} (a), the denoising block of the diffusion transformer utilizes a self-attention mechanism~\cite{esser2024scaling} that early concatenates image and text embeddings, allowing them to be jointly flowed into self-attention blocks to predict the noise. In contrast, as demonstrated in Fig~\ref{fig:attention} (b), the UNet architecture employs cross-attention layers to introduce text conditions into the denoising process, which might learn entangled mapping from input conditions to image semantics~\cite{liu2024towards}. For example, when generating images with UNet-based diffusion models using the caption \textit{"a $<$color$>$ $<$object$>$"}, the attention map w.r.t. semantic "$<$color$>$" might contain category information of semantic "$<$object$>$". Therefore, when we aim to edit semantic "$<$color$>$", the semantic "$<$object$>$" would be incorrectly modified. 

Building on this insight, to understand the origin of the disentanglement effect, we conduct a probing analysis~\cite{clark2019does,liu2019linguistic} to explore the differences between the self-attention and cross-attention mechanisms in guiding image generation, as utilized in transformer-based and UNet-based diffusion models, respectively. %Without loss of generality, we analyze the attention maps of text tokens in our experiments. 
Specifically, we generate images using the caption "a $<$color$>$ $<$object$>$" and analyze the attention maps corresponding to each token in the intermediate layers. We then train classifiers using the attention maps of "$<$color$>$" and apply these classifiers to the attention maps of "$<$object$>$. If classifiers successfully identify color information within the "$<$object$>$" attention maps, the attention map may contain category information related to "$<$color$>$". 

\begin{figure}
    \centering
    \includegraphics[width=1\linewidth]{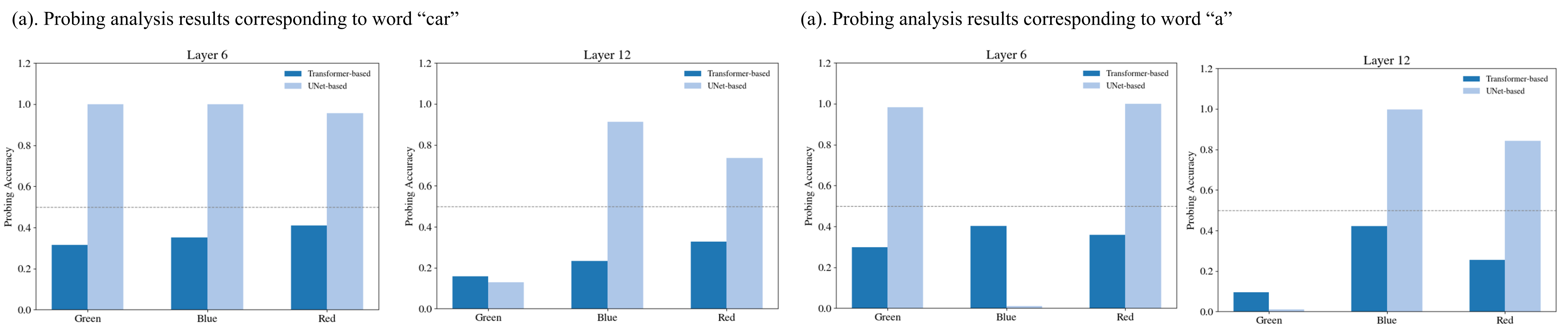}
    \caption{Results of the probing analysis. We provide ratios of attention maps from different models being classified into various color categories. when tested against corresponding color classifiers.}
    \label{fig:linear prob}
    \vspace{-15pt}
\end{figure}

\begin{figure*}[h!] 
    \centering
    
    \includegraphics[width=1\textwidth]{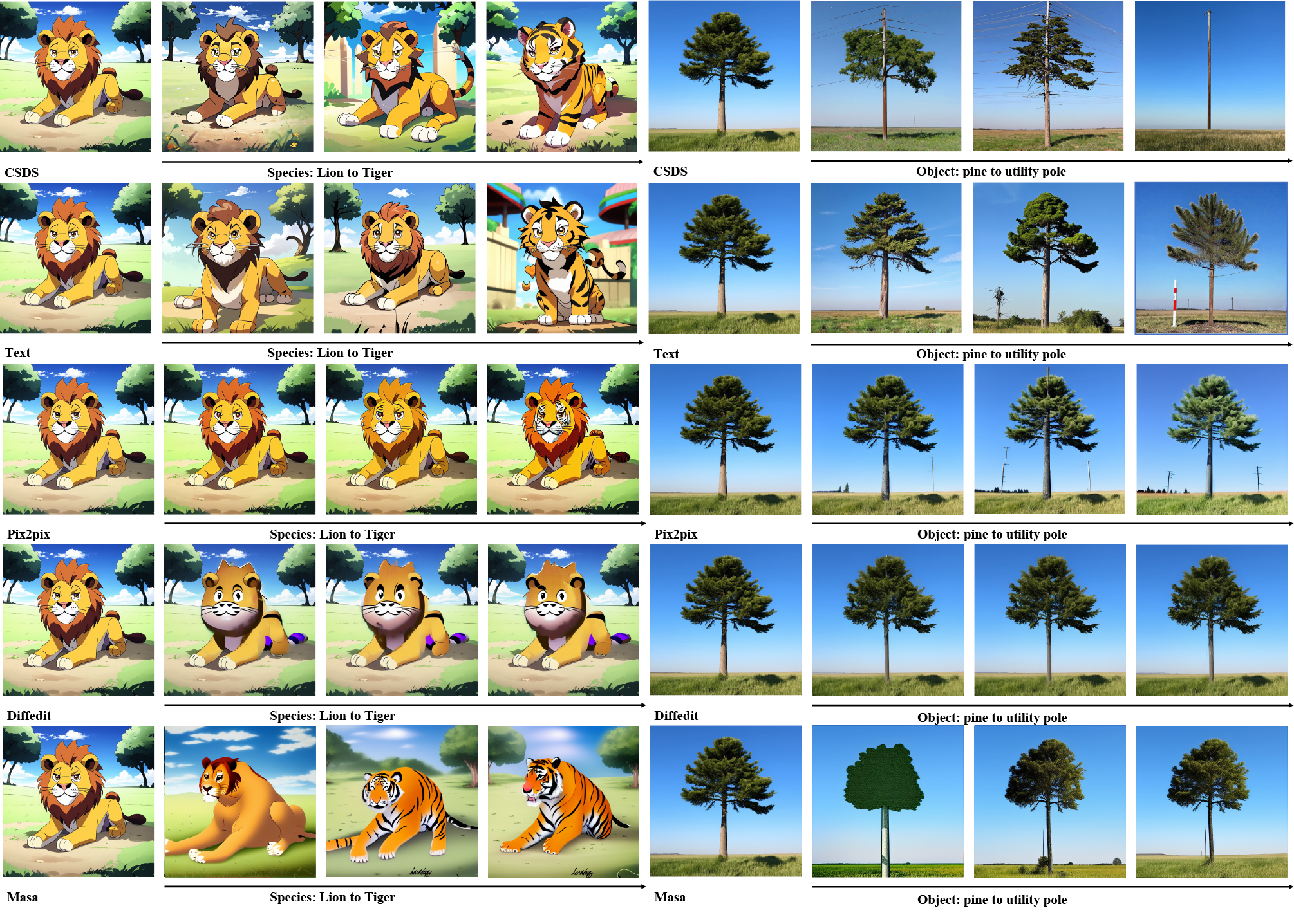} % Adjust the path and width as needed
    %\vspace{-10pt}
    \caption{Comparisons of fine-grained editing results of our EIM method and baseline methods. Our EIM method has successfully achieved progressive transformations from lion to tiger and pine tree to utility pole. While baseline methods fail to achieve the target transformations, often degrading the original image quality significantly or producing unintended effects.
    }
    \vspace{-10pt}
\label{baseline}
\end{figure*}

\textbf{The attention map for a specific semantic in the diffusion transformer does not contain category information from other semantics.} As illustrated in Fig.~\ref{fig:linear prob}, the ratio of the diffusion transformer's attention map for ``$<$car$>$'' being classified with a specific color is near 0.5, indicating that it does not effectively encode color category information. This contrasts with the attention map of ``$<$car$>$'' in UNet-based diffusion models, which shows a higher tendency in identifying color categories. This observation suggests that the disentanglement property in diffusion transformers stems from their image-text joint self-attention mechanism, leading to less entanglement in the representations corresponding to each semantic. Which similar to the findings in~\cite{zhang2024cross}. Further experimental details and results are provided in Appendix~\ref{appendix analysis study}.

\vspace{-10pt}
\subsection{Bidirectional Editing and Limitations}
\label{boundary and bidirection}
When the latent vector moves across the hyperplane formed by the editing direction based on the target value of the semantic (e.g., semantic value "smile" of the semantic "expression", as discussed in Sec.~\ref{disentangle space sec}), the sign of the distance between the latent representation and the hyperplane flips. This allows us to both increase and decrease the intensity of a given semantic value.

Formally, given a normal vector $n_i$ of editing direction for semantic $s_i,i\in[1,...,m],m\in N^+$, and a latent representation $z$, the signed distance to the hyperplane is given by: $D = n_i^T z$. As $z$ moves in the direction of $n_i$ or $-n_i$, the distance $d$ becomes positive or negative, respectively, allowing bidirectional control of the attribute.

\begin{figure*}[htp] 
    \centering
    
    \includegraphics[width=1\textwidth]{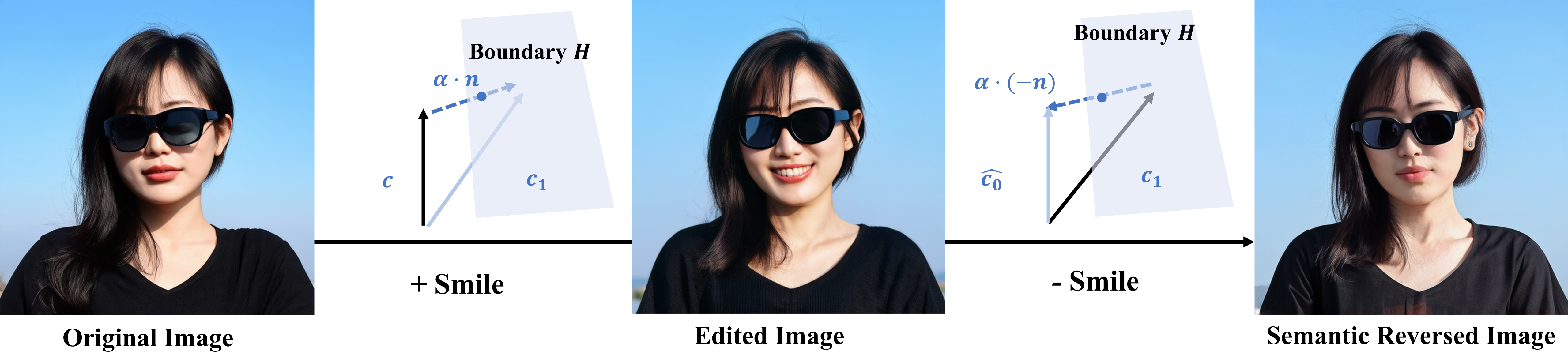} % Adjust the path and width as needed
    %\vspace{-10pt}
    \caption{Edited semantics can be reverted to their original values. The disentanglement property allows precise editing of the target semantic without affecting others, enabling us to use the same direction that was used to edit the semantic to reverse it back to its original value.
    }
    \vspace{-15pt}
\label{ablation_reverse}
\end{figure*}

As shown in Fig.~\ref{ablation_reverse}, we have achieved bi-directional image editing across the boundary defined by the editing direction. We first manipulate the latent embedding along the direction given by our method to obtain the edited image. Then, the latent embedding is manipulated in the opposite direction to obtain the semantic reversed image. As shown in the figure, the expression of the woman is first changed to "smile" and then changed from "smile" to "serious".

\begin{figure}
    \centering
    \includegraphics[width=\linewidth]{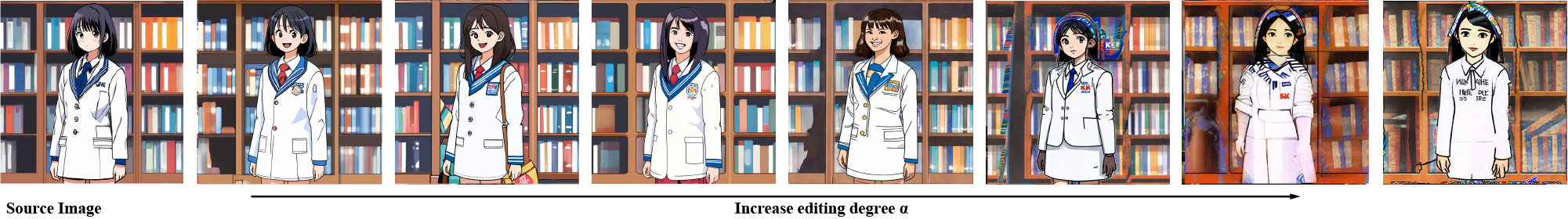}
    \caption{Linear manipulation is effective near the boundary defined by the editing direction. However, if the edit degree exceeds the threshold set by Proposition.(\ref{threshold}), the resulting image may become corrupted.}
    \label{fig:corrupt}
\end{figure}

We also find that effective manipulations with $n$ should be near the boundary defined by it. In other words, the editing degree $\alpha$ could not be extremely large. As demonstrated in Fig.~\ref{ablation_reverse}, when we move the latent vector $z$ too far from the hyperplane, exceeding a certain threshold, the probability of the projection falling near the hyperplane diminishes rapidly. This is consistent with the concentration of measure phenomenon in high-dimensional spaces \cite{shen2020interpreting}. In our work, the threshold can be given by Proposition~\ref{threshold} in Sec.~\ref{disentangle space sec}. When the threshold is exceeded, we observe a loss of controlled attribute editing. For example, in our experiments with facial attributes, we find that extreme modifications to the "expression" semantic begin to affect unrelated features such as ethnicity, as shown in Fig.~\ref{fig:corrupt}.

%Mathematically, this can be expressed as:

%\begin{equation}
%P(|n_i^T z| \leq \epsilon) \geq 1 - 2e^{-\epsilon^2/2}
%\end{equation}

%where $\epsilon$ is the threshold distance from the hyperplane.

%This phenomenon highlights the importance of careful calibration of the editing strength $\alpha$ (or the diagonal elements of $\Lambda$ in multi-attribute editing) to maintain precise control over the desired attributes while preserving other semantic features of the image.

\subsection{Hyperparameter Analysis}

\begin{figure*}[htp] 
    \centering
    
    \includegraphics[width=1\textwidth]{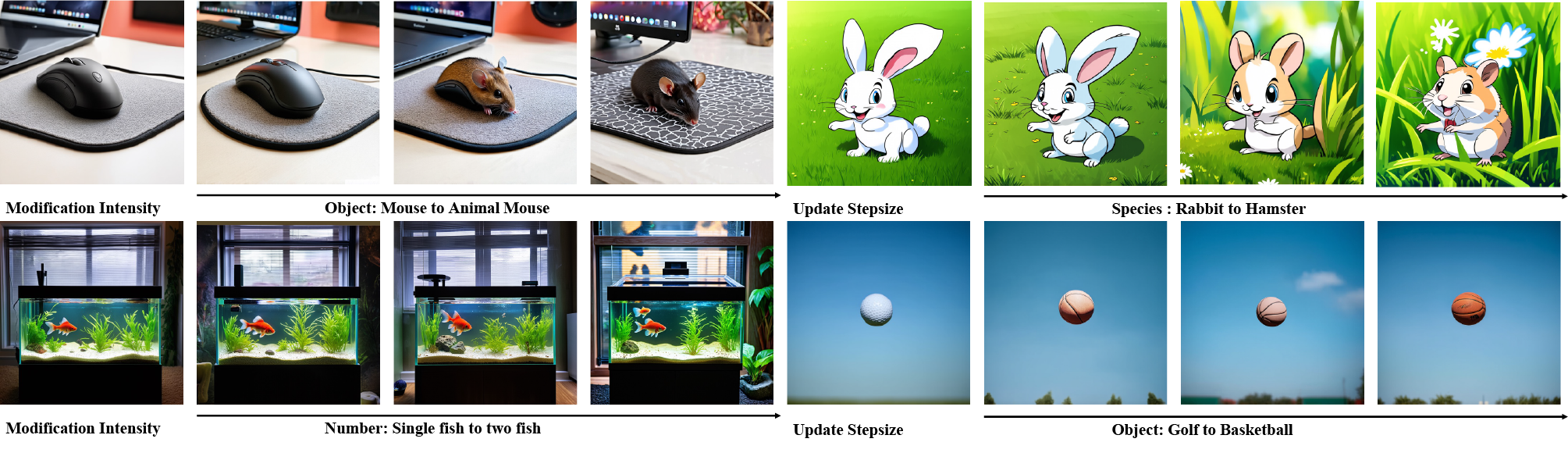} % Adjust the path and width as needed
    %\vspace{-10pt}
    \caption{Ablation study on hyperparameters in our EIM method. The figure demonstrates the effects of modification intensity ($\lambda$) and update stepsize on various editing tasks. These examples highlight how careful tuning of these hyperparameters allows for precise control over edit strength, content preservation, and the balance between desired modifications and maintaining original image semantics.
    }
    \vspace{-15pt}
\label{hyper-param 1}
\end{figure*}

In addition to the editing strength $\alpha$, two crucial hyperparameters in our EIM method are the update stepsize and the modification intensity $\lambda$ used in the Hessian score-distillation sampling (HSDS) process.

\textbf{Update Stepsize:} The update stepsize $\eta$ determines the magnitude of each update during the Score-Distillation Sampling process. We find that a carefully chosen stepsize is essential for achieving smooth and controlled edits. Too large a stepsize can lead to overshooting and instability in the editing process, while too small a stepsize can result in slow convergence and subtle edits.

In our experiments, we employ an adaptive stepsize strategy, starting with a relatively large value of 0.1 and gradually decreasing it to 0.01 as it progresses. This approach allows for rapid initial progress followed by finer adjustments, resulting in more precise edits.

\textbf{Modification Intensity $\lambda$:} The modification intensity $\lambda$ in our HSDS objective (Eq.~\ref{csds}) plays a crucial role in balancing between achieving the desired edit and preserving the original image content. It controls the trade-off between modifying the target attribute and maintaining other semantic features of the image. We empirically observe that setting $\lambda$ in the range of 0.1 to 1.0 generally produces good results across various editing tasks. Specifically, a higher $\lambda$ values (0.7 - 1.0) result in more conservative edits that better preserve the original image structure but may require more iterations to achieve the desired effect. A Lower $\lambda$ values (0.1 - 0.3) allow for more aggressive edits but may occasionally introduce unintended changes in other attributes. The optimal value of $\lambda$ can vary depending on the specific attribute being edited and the desired strength of the edit. For example, we observe that edits involving more localized changes (e.g. facial features) generally benefit from higher $\lambda$ values, while global edits (e.g. overall style or lighting) can tolerate lower $\lambda$ values. Figure~\ref{hyper-param 1} illustrates the effect of different $\lambda$ values on the editing process for a sample image, demonstrating the trade-off between edit strength and content preservation.

\begin{figure}
    \centering
    \includegraphics[width=\linewidth]{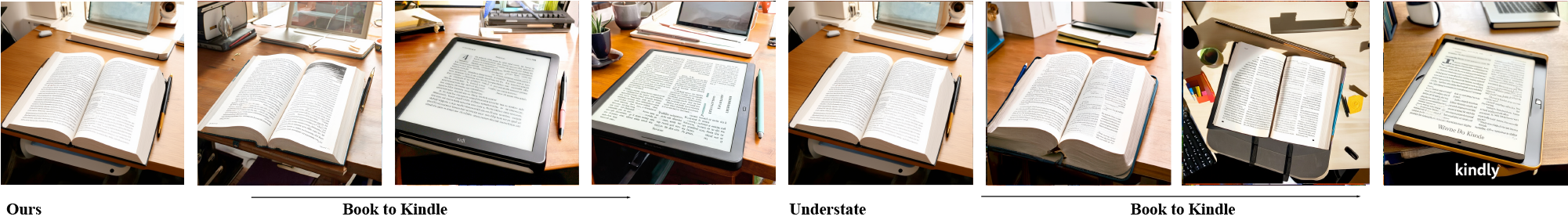}
    \caption{Highlight editing targets in the HSDS can help manipulate image embedding more effectively.}
    \label{fig:csds add}
\end{figure}

\textbf{Score-Distillation-Sampling (SDS) guided by text prompt of the editing target can provide a more effective editing direction.} We use the text embedding encoded from the text prompt of the given editing target, to guide the naive SDS process to provide an editing direction. As shown in Fig.~\ref{fig:csds add}, compared to utilizing the text description of the source image to guide the SDS process, this approach leads to more effective image manipulation, but results in more unintended changes. Therefore, use text prompt of the editing target in SDS can provide a more effective but less precise editing direction. This finding has motivated our HSDS method as discussed in the appendix~\ref{sec theory HSDS}.

%These findings not only provide insight into the behavior of our EIS method but also shed light on the structure of the latent space in diffusion transformers. Understanding these properties and limitations is crucial for developing more robust and controllable image editing techniques in future work. Further research could explore automatic adaptation of these hyperparameters based on the specific editing task and image content, potentially leading to more robust and user-friendly image editing systems.

\subsection{Detailed Analysis on Semantic Loss}

\begin{figure*}[htp] 
    \centering
    
    \includegraphics[width=1\textwidth]{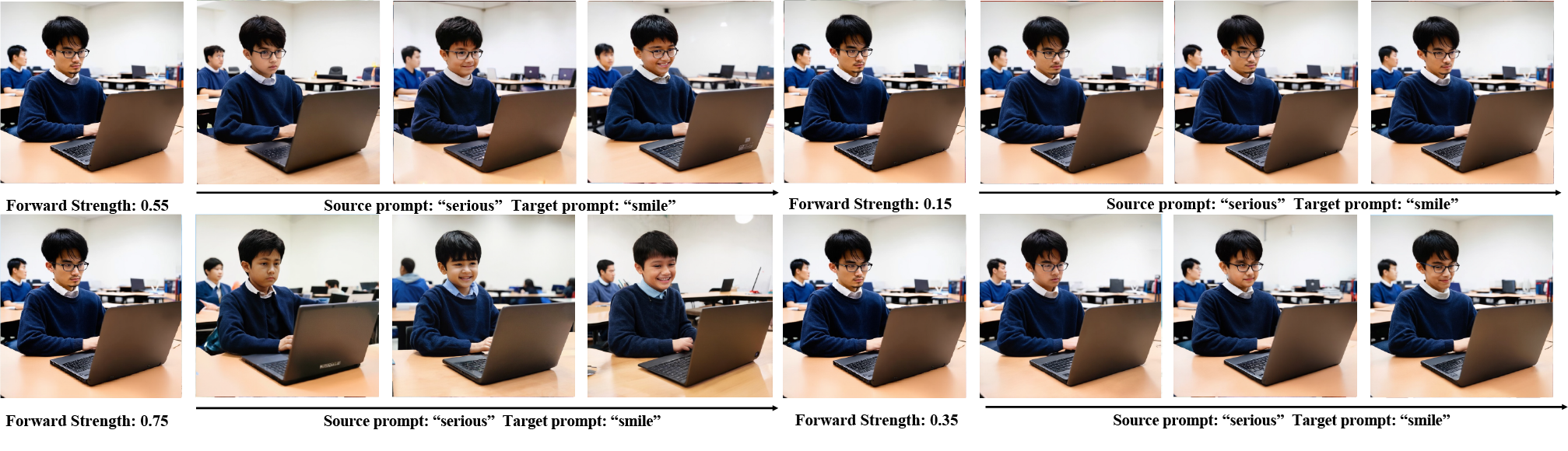} % Adjust the path and width as needed
    %\vspace{-10pt}
    \caption{Semantics are gradually lost during the forward process. If not conditioned, these lost semantics are randomly recovered. As the forward strength increases, semantics such as 'age,' 'ethnicity,' and 'glasses' are sequentially lost, and in the recovered images, these semantics are assigned random values.
    }
    %\vspace{-10pt}
\label{ana1}
\end{figure*}

\textbf{In image editing, semantics are gradually lost as the forward strength increases.} 
We forward the given image with 4 different strengths, resulting in 4 sets of noised images with different noise degrees. We de-noise the images conditioned on the editing target, and obtain 4 sets of recovered images. As shown in Fig.~\ref{ana1}, at lower forward strengths (e.g., 0.15), the model successfully preserves the target semantic "smile" while keeping other attributes, such as "ethnicity," unchanged. However, the semantic "smile" has not been significantly changed when forward strength is 0.15. Meanwhile, as the forward strength increases to 0.55 and beyond, critical semantics like "ethnicity," "glasses," and "age" are replaced with random values in the reconstructed images. This progression demonstrates the trade-off required in controlling forward strength to prevent unintended alterations in semantics irrelevant to editing tasks.

\begin{table}[h] % The 't' means "top of the page"
\centering % This centers the table

\caption{Comparisons between related works and our paper. We organize related works according to our key areas of interest. For properties of the considered semantic latent space, we focus on: (1) Semantic latent space is easy to identify. (2) Image semantics are not entangled. (3) Linear manipulation of the semantic space can linearly modify the intensity of semantics. For the image editing application, we focus on: (1) Precisely edit intensity of semantics. (2) Zero-shot image editing. (3) No need for reference images.}
\scalebox{0.9}{
\begin{tabular}{@{} l lll lll }
\toprule
\textbf{Method} 
& \textbf{Easy Identify} &\textbf{Disentangled} & \textbf{Linearity}& \textbf{Precise}& \textbf{Zero-Shot} & \textbf{Reference-Free}  \\
%\cmidrule(l){2-4} \cmidrule(l){5-7} \cmidrule(l){8-10} \cmidrule(l){11-13} 
\midrule
~\cite{hertz2022prompt}& & & & & $\checkmark$& \\
~\cite{cao2023masactrl}& & & & & $\checkmark$& \\
~\cite{brooks2023instructpix2pix}& & & & & &$\checkmark$ \\
\midrule
~\cite{couairon2022diffedit}& & & & &$\checkmark$ & $\checkmark$ \\
~\cite{hertz2023delta}& & & & & $\checkmark$ & $\checkmark$ \\
\midrule
~\cite{wu2023uncovering}& $\checkmark$& & & & &$\checkmark$ \\
~\cite{kawar2023imagic} & & $\checkmark$ & & $\checkmark$ & & $\checkmark$ \\
\midrule
EIM (ours) & $\checkmark$ & $\checkmark$ & $\checkmark$ & $\checkmark$ & $\checkmark$ & $\checkmark$ \\
\bottomrule
\end{tabular}
}
\label{table1}
\vspace{-10pt}
\end{table}
\section{Related Work}
\vspace{-5pt}
\subsection{Diffusion Model}
In recent years, diffusion models have solidified their position as leading techniques in image generation, particularly for producing high-quality and diverse visuals~\cite{ho2020denoising, rombach2022high, esser2024scaling}. These models traditionally introduce noise to images in a forward process and utilize a U-Net architecture~\cite{ronneberger2015u} to predict the denoised outputs. Text-to-image diffusion models have particularly stood out, surpassing traditional generative models like GANs~\cite{karras2019style} in both image quality and zero-shot synthesis capabilities~\cite{rombach2022high, ramesh2022hierarchical}. While U-Net has been the standard architecture for denoising, recent research has shifted towards transformer-based diffusion models~\cite{peebles2023scalable, esser2024scaling}, demonstrating superior performance in generating high-fidelity and diverse images. Despite the successes of diffusion transformers, the precise mechanisms by which input conditions influence the generation process remain largely unexplored.
\vspace{-5pt}
\subsection{Semantic Disentanglement in Generative Model}
Semantic disentanglement is an essential property in previous generative models, like Generative Adversarial Networks~\cite{goodfellow2020generative, karras2019style} and $\beta$-Variational Autoencoder~\cite{burgess2018understanding}, enabling precise control over how specific semantics are represented in the generated samples. With a generative model that has a well-disentangled latent space, we can modify the values of specific semantics in the latent space without affecting others, thereby achieving precise and precise image editing~\cite{shen2020interpreting}. %Previous studies have demonstrated that a disentangled latent space exists within GAN-based models, especially in the StyleGAN~\cite{karras2019style}. 
By adjusting specific directions in this latent space, it is possible to modify only the targeted semantics~\cite{harkonen2020ganspace}. While these models exhibit decent results in continuously and precisely modifying specific semantics, their generated samples are restricted in the distribution of their training datasets, and thus lack zero-shot generating ability~\cite{dhariwal2021diffusion}. On the other hand, diffusion models~\cite{ho2020denoising}, especially text-to-image diffusion models~\cite{saharia2022photorealistic, rombach2022high}, are remarkable in synthesizing diverse and high-fidelity samples~\cite{ho2022cascaded} and composing unseen images~\cite{liu2022compositional}. ~\cite{hertz2022prompt, kwon2022diffusion, wu2023uncovering} have demonstrated that diffusion models possess a semantic latent space with word-to-semantic mappings, this property has been utilized for attention-map-based image editing and generation~\cite{cao2023masactrl,chefer2023attend,chen2024training}. Other related works~\cite{wu2023not,wang2023stylediffusion} have considered the disentanglement in the latent space of the text embedding space, or explored the feasibility of seeking disentangled latent space~\cite{lu2024hierarchical, everaert2023diffusion, everaert2024exploiting,hahm2024isometric}. However, the editing directions for precise and precise changing desired semantics are not explicit, posing a non-trivial challenge in manipulating semantics in the latent space.

\subsection{Image Editing with Text-to-Image Diffusion Model}
Recently, the capability of image editing of UNet-based Text-to-Image (T2I) diffusion models has been widely studied. Inspired by the semantic meaning of image-text attention maps, ~\cite{hertz2022prompt} demonstrates a word-to-semantic relationship in UNet-based T2I diffusion models, which motivates a series of image-editing methods based on manipulating attention maps~\cite{brooks2023instructpix2pix,cao2023masactrl,chefer2023attend}. Due to the entangled latent space of UNet-based diffusion model, these approaches often can only achieve coarse grained editing. ~\cite{kawar2023imagic,wu2023uncovering} proposes to optimize image embedding to achieve disentangled editing. However, these approaches are not zero-shot and require extra fine-tuning on model parameters. To alleviate the entangled correlation between text tokens and image semantics, another type of methods~\cite{hertz2023delta,nam2024contrastive} consider utilize score-distillation-sampling-based methods to iteratively update the noised image latents for precisely editing target items. While these works have improved the ability of T2I diffusion models to follow editing instructions, leveraging them for text-based precise image editing remains challenging. Our work verifies the disentanglement in the representation space of the diffusion transformer and thus enables precise image editing via latent space manipulation.

\vspace{-5pt}
\section{Conclusion}
In this work, we explored the semantic disentanglement properties of the latent space of Text-to-Image diffusion transformers. Through extensive empirical analysis, we identified that the latent space of text-to-image DiTs is inherently disentangled, enabling precise and precise control over image semantics. Based on these insights, we proposed a novel Encode-Identify-Manipulate (EIM) framework that leverages this disentanglement property for zero-shot precise image editing. Our method demonstrates significant advancements over existing approaches, providing a robust and efficient solution for precise and precise image editing without additional training. Furthermore, we introduced a benchmark and a metric to quantitatively assess the disentanglement properties of generative models. We believe our work will inspire further research and serve as a valuable resource for advancing studies in generative models.

\clearpage
\appendix
%%%%%%%%%%%%%%%%%%%%%%%%%%%%%%%%%%%%%%%%%%%%%%%%%%%%%%%%%%%%%%%%%
\section{Theoretical Analysis}
\subsection{Formal Modeling of Disentanglement Properties}
\label{sec formal disentanglement}
\label{def disentangle}
In the classic modeling of visual generation and feature representation~\cite{wang2021self, higgins2018towards}, let $\mathcal{S}$ represents a set of unseen semantics (e.g., "color" and "style"), where each semantic can take on various values (e.g., color: "purple" or "red"). There exists an generation process $\Phi: \mathcal{S} \rightarrow \mathcal{I}$ that generates images from semantics (e.g., thinking about a red dress $\rightarrow$ painting a red dress). Ideally, $\mathcal{S}$ can be decomposed into the Cartesian product of $m$ modular semantic subspaces $\mathcal{S} = \mathcal{S}_1 \times \ldots \times \mathcal{S}_m, m \in \mathbb{N}^+$~\cite{yue2024exploring}. With this property, we can locally intervene the modular semantic $s_i \in \mathcal{S}_i, i \in [1,\ldots,m]$ without affecting the others (e.g., in facial images, changing the value of semantic "glasses" will not modify "ethnicity" and "expression"). Then we can consider all image editing operations (e.g., changing semantic "expression") as a group $G$ acting on $\mathcal{S}$, which can be decomposed as the direct product of them: $G = g_1 \times \ldots \times g_m$. The intervention on semantic $s_i$ can be defined as the operator $g_i: \mathcal{S}_i \rightarrow \mathcal{S}_i$, which manipulates $s_i$ by $\tilde{s_i} = g_i(s_i)$. 

However, $\mathcal{S}$ is an ideal, unobservable space and therefore cannot be directly leveraged for downstream generation tasks. In visual representation learning, we typically approximate $\mathcal{S}$ by learning a semantic representation space $\mathcal{S}'$ through a neural network $f: \mathcal{I} \rightarrow \mathcal{S}'$. In this work, we consider the joint latent space $\mathcal{Z}$ discussed in Sec.~\ref{latent space}, is this semantic representation space $\mathcal{S}'$. During denoising loops, $z_t$ and $c$ are first passed into the denoising transformer to obtain $z_0$, then they are decoded back to an image in $\mathcal{I}$. We define this process of denoising and decoding as function $h: \mathcal{S}' \rightarrow \mathcal{I}$.

In image editing, the joint latent embeddings are first modified to obtain \( (\tilde{z_t}; \tilde{c}) \). They are then mapped by \( h \) to produce the edited image. However, the intervention on semantic representation space $\mathcal{S}'$ might be inaccurate and not controllable. To achieve precise image editing, semantics should be encoded into different subspaces of $\mathcal{S}'$, where they are controlled by separate editing directions. This requires $\mathcal{S}'$ to be a disentangled semantic representation space, which should satisfy the following properties:

\begin{property}
\label{remark decompose}
\textbf{Decomposable:} For a decomposable semantic representation space $S'$, there exists a decomposition\footnote{Here in the decomposability, we have not assumed that semantics in training datasets or testing dataset are statistically independent. We focus on the property that "modifying desired semantic without changing others".} such that $S' = S'_1 \times \cdots \times S'_m$ corresponding to semantics in the real semantic space $S$. Therefore, $S'_i$ is only affected by $g_i$. For example, changing the "expression" vector in $S'$ does not affect the "ethnicity" semantic in the edited image.
\end{property}

\begin{property}
\label{remark effectiveness}
\textbf{Effectiveness:} For any editing operation $g_i \in \mathcal{G}$, and any pair of latent representation semantic and its corresponding real-world semantic $s_i \in \mathcal{S}$, there exists $h(g_i(s_i')) = \phi(g_i(s_i))$. For example, changing the intensity of semantic "smile" in the unseen $S'$ will be equivalent to directly changing the real image semantic in $\mathcal{S}$ with the direction of "smile".
\end{property}

The decomposability can be empirically verified by checking if \( S'_i \) is only affected by \( g_i \) for semantic \( s_i \). The second property can be validated by systematically testing whether different semantics can be effectively edited.

\subsection{Feasibility of linearly manipulating semantics in the latent space}
\label{sec:feasibility}
The disentanglement properties of the diffusion transformer not only enable precise semantic editing, but also enable linearly adjusting the editing degree $\alpha$ that a semantic moves toward a specific value. For a given semantic \( s \) and the corresponding editing direction $n\in \mathbb{R}^{md}$, which defines a hyperplane boundary \( H \subset \mathbb{R}^{md} \) given by \( H = \{ h \in \mathbb{R}^{md} : h \cdot n = 0 \} \), linear manipulation could be performed near $H$, and achieve precise control~\cite{shen2020interpreting}. 

Specifically, in text-to-image diffusion models, text prompts are encoded to token-type embeddings as illustrated in Sec.~\ref{Preliminary}, where each semantic is encoded to different subspaces $C_i\in \mathbb{R}^d,i\in[1,...,m]$.\footnote{We assume each text token corresponds to a specific semantic, and subspaces in \( c_1, c_2, \dots \) correspond to semantics in \( S_1, S_2, \dots \) in a 1-to-1 mapping. Without loss of generality, we disregard the possibility that the position of each semantic's text token may vary across different prompts.} Then we can obtain editing directions $n_i$ of them for manipulations of the text representation on these subspaces. The feasibility of manipulations near the boundary hyper-plane is guaranteed by the following proposition:

\begin{proposition}
\label{threshold}
Let the editing directions $\mathbf{n}_1, \mathbf{n}_2, \dots, \mathbf{n}_m \in \mathbb{R}^d$ be unit vectors, i.e., $\|\mathbf{n}_i\| = 1$ for all $i = 1, \dots, m$. Define $\mathbf{n}_i^{\text{ext}} \in \mathbb{R}^{md}$ as the extension of $\mathbf{n}_i$ into $\mathbb{R}^{md}$ by placing $\mathbf{n}_i$ in the $i$-th block of $d$ coordinates and filling the rest with zeros. Let $\mathbf{z} \sim \mathcal{N}(0, \mathbf{I}_{md})$ be a multivariate random variable. Then we have:
\[
\mathbb{P}\left( \left|\sum_{i=1}^{m} \mathbf{n}_i^T \mathbf{z}_i \right| \leq 2\alpha \sqrt{\frac{d}{d - 2}} \right) \geq \left( \left( 1 - 3e^{-cd} \right) \left( 1 - \frac{2}{\alpha} e^{-\alpha^2 / 2} \right) \right)^m,
\]
for any $\alpha \geq 1$ and $d \geq 4$. Here, $\mathbb{P}(\cdot)$ stands for probability and $c$ is a fixed positive constant.
\end{proposition}

This proposition establishes a bound on how close the vector should be to the hyper-plane. It's important to note that the editing degree should not be excessively large, as increasing it too much, as discussed in Sec.~\ref{boundary and bidirection}, can lead to incorrect image generation or even cause the model to collapse.
%%%%%%%%%%%%%%%%%%%%%%%%%%%%%%%%%%%%%%%%%%%%%%%%%%%%%%%%%
\subsection{Extension to Multi-Attribute Editing}
\label{sec:multi-attr edit}
Our EIM method extends to multi-attribute editing scenarios. Formally, let a semantic \( s \in \mathcal{S} \) comprise modular semantics \( s_i \in \mathcal{S}_i \) for \( i \in [1, \dots, m] \). The editing direction \( n_i \) for each \( s_i \) is derived by encoding a text prompt for the target value of \( s_i \). Extending \( n_i \) to the full space, we obtain \( \mathbf{n}_i^{\text{ext}} \in \mathbb{R}^{md} \). These extended directions are mutually orthogonal, enabling precise and multi-attribute semantic editing.

\begin{proposition}
    Let the editing directions $\mathbf{n}_1, \mathbf{n}_2, \dots, \mathbf{n}_m \in \mathcal{C}_1,...,\mathcal{C}_m$ of semantics be unit vectors, i.e., $\|\mathbf{n}_i\| = 1$ for all $i = 1, \dots, m$. Define $\mathbf{n}_i^{\text{ext}} \in \mathbb{R}^{md}$ as $\mathbf{n}_i^{\text{ext}} = (0_d, \ldots, 0_d, \mathbf{n}_i, 0_d, \ldots, 0)$, where $\mathbf{n}_i$ occupies the $i$-th block of $d$ coordinates. Then, $\{\mathbf{n}_i^{\text{ext}}\}_{i=1}^{m}$ forms an orthogonal set in $\mathbb{R}^{md}$, i.e., for $i \neq j$, $\mathbf{n}_i^{\text{ext}} \cdot \mathbf{n}_j^{\text{ext}} = 0$.    
\end{proposition}

\subsection{Discussion on How HSDS Push the Image Embedding to Target Direction}
\label{sec theory HSDS}
We aim to seek an editing direction $n$ in the joint latent space that controls the target semantic that will be edited. Specifically, the editing direction on the text subspace can be easily identified, and here we focus on obtaining the direction in the image subspace with a Score-Distillation-Sampling (SDS) based approach. We are given a input image (or image embedding encoded from it) $z$ and a text embedding $c_1$ encoded from a text prompt that describes the desired image, a text embedding $c_0$ encoded from a text prompt that describes the original image, as well as a text prompt $\text{attr}$ of the desired value of the target semantic.

In image editing task, the SDS can provide an direction $\nabla \mathcal{L}_{\text{SDS}}$ through differentiating the noise-prediction loss. Here the predicted loss is given by a frozen diffusion model. However, in our text-guided image editing task, directly applying SDS may cause modifications that are irrelevant to input text conditions as discussed in~\citep{hertz2023delta}. To address this challenge, previous work~\citep{hertz2023delta} decomposes the gradients $\nabla_\theta \mathcal{L}_{\text{SDS}}$ to two components:
\[
\nabla_\theta \mathcal{L}_{\text{SDS}}(z, c_1, \epsilon, t) := \delta_{\text{text}} + \delta_{\text{bias}},
\]
where $\epsilon$ is the noise and $t$ is the timestep, $\delta_{\text{text}}$ directs the image towards the closest match to the text embedding, while the undesired component $\delta_{\text{bias}}$ introduces smoothness and blurriness. And the text-aligned part $\delta_{\text{text}}$ can be viewed as the editing direction induced by $c_1$. Notably, $\theta$ is the parameter of an image render \( g_{\theta} \). In the following sections, we assumes \( g_{\theta} = \theta\).

Based on this intuition, ~\citep{hertz2023delta} has proposed a Delta-Denoising-Score to obtain the direction $\delta_{\text{text}}$ through a iterative loop, based on the following equation:
\[
\mathcal{L}_{\text{DDS}}(z, c_0, z', c_1, \epsilon, t) = \left\| \epsilon_{\phi}(z_t, c_0, t) - \epsilon_\phi(z_t', c_1, t) \right\|_2^2,
\]
where $\phi$ is the de-noising block. $\hat{z}$ is the intermidiate embedding in the iterative loop, which is updated following $z' = z' - \eta \nabla_\theta \mathcal{L}_{\text{DDS}}$, where $\eta$ is the update stepsize. Then, the gradient over $\theta$ of $z$ are given by
\[
\nabla_\theta \mathcal{L}_{\text{DDS}} = \left( \epsilon_\phi(z_t, c_0, t) - \epsilon_\phi(z_t', c_1, t) \right) \frac{\partial z}{\partial \theta}.
\]

Based on analysis in~\citep{hertz2023delta}, we can represent DDS as a difference between two SDS scores:
\[
\nabla_\theta \mathcal{L}_{\text{DDS}} = \nabla_\theta \mathcal{L}_{\text{SDS}}(z, c_0) - \nabla_\theta \mathcal{L}_{\text{SDS}}(z', c_1).
\]

Although DDS provides an editing direction, \(\nabla_\theta \mathcal{L}_{\text{DDS}}(z, c_1, c_0, \epsilon, t)\), that aligns with the text condition \(c_1\), it cannot offer a disentangled direction to control specific target semantics. An alternative is to use \(\nabla_\theta \mathcal{L}_{\text{DDS}}(z, c_{\text{attr}}, c_0, \epsilon, t)\) as the editing direction, where \(z_{\text{attr}}\) represents the text embedding derived from a prompt describing the desired target semantic. However, this approach may align solely with \(z_{\text{attr}}\) and overlook task-irrelevant context information in \(c_1\), as discussed above.

Therefore, we divide the DDS term into two parts:
\[
\nabla_\theta \mathcal{L}_{\text{HSDS}}(z, c_1, c_0, \epsilon, t) := \delta_{\text{attr}} + \delta_{\text{context}},
\]
where $\delta_{\text{attr}}$ represents the editing direction of the target semantic, given the context text embedding $c_1$ which describes the desired image. And $\delta_{\text{context}}$ represents the direction of the context part. 

Therefore, we aim to utilize the direction given by $\delta_{\text{attr}}$ for precise image editing. We construct the Hessian Score-Distillation-Sampling by: 
\[
\nabla_\theta \mathcal{L}_{\text{HSDS}} = \nabla_\theta \mathcal{L}_{\text{DDS}}(z, c_{attr}, c_0) - \nabla_\theta \mathcal{L}_{\text{DDS}}(z, c_1, c_0).
\]

It can be rewritten as:
\[
\mathcal{L}_{\text{HSDS}}(z, c_0, z', c_1, c_{\text{attr}}, \epsilon, t) = \left\| \epsilon_{\phi}(z_t, c_{\text{attr}}, t) - \epsilon_\phi(z_t', c_0, t) \right\|_2^2 - \left\| \epsilon_{\phi}(z_t, c_1, t) - \epsilon_\phi(z_t', c_0, t) \right\|_2^2,
\]
and we can differentiate it to obtain the editing direction.

\section{Proof}
In the section, we provide proofs for the theorem proposed in Sec.~\ref{sec:feasibility} and Sec.~\ref{sec:multi-attr edit}.

\subsection{Proof for Proposition 1}
\begin{proof}

   Let $\mathbf{n}_1, \mathbf{n}_2, \dots, \mathbf{n}_m \in \mathbb{R}^d$ be $m$ unit vectors such that $\|\mathbf{n}_i\| = 1$ for all $i = 1, \dots, m$. Each $\mathbf{n}_i$ can be extended to $\mathbf{n}_i^{\text{ext}} \in \mathbb{R}^{md}$ by placing $\mathbf{n}_i$ in the $i$-th block of $d$ dimensions and filling the remaining entries with zeros. Thus,
\[
\mathbf{n}_i^{\text{ext}} = (0, \dots, 0, \mathbf{n}_i, 0, \dots, 0),
\]
where $\mathbf{n}_i$ occupies the $i$-th block of $d$ coordinates.

Let $\mathbf{z} \sim \mathcal{N}(0, \mathbf{I}_{md})$ be a standard Gaussian random vector in $\mathbb{R}^{md}$. We want to prove the inequality:
\[
\mathbb{P}\left( \left|\sum_{i=1}^{m} \mathbf{n}_i^T \mathbf{z}_i \right| \leq 2\alpha \sqrt{\frac{d}{d - 2}} \right) \geq \left( \left( 1 - 3e^{-cd} \right) \left( 1 - \frac{2}{\alpha} e^{-\alpha^2 / 2} \right) \right)^m.
\]

We start by expressing the inner product $\mathbf{n}^T \mathbf{z}$ as a sum of independent terms:
\[
\mathbf{n}^T \mathbf{z} = \sum_{i=1}^{m} (\mathbf{n}_i^{\text{ext}})^T \mathbf{z}.
\]
Each term $(\mathbf{n}_i^{\text{ext}})^T \mathbf{z}$ is equivalent to $\mathbf{n}_i^T \mathbf{z}_i$, where $\mathbf{z}_i$ is the $i$-th block of $d$ dimensions in $\mathbf{z}$. Therefore,
\[
\mathbf{n}^T \mathbf{z} = \sum_{i=1}^{m} \mathbf{n}_i^T \mathbf{z}_i.
\]

Using Property 2 in~\cite{shen2020interpreting}, for each $\mathbf{n}_i \in \mathbb{R}^d$ and $\mathbf{z}_i \sim \mathcal{N}(0, \mathbf{I}_d)$, we have:
\[
\mathbb{P}\left(|\mathbf{n}_i^T \mathbf{z}_i| \leq 2\alpha \sqrt{\frac{d}{d - 2}}\right) \geq \left(1 - 3e^{-cd}\right)\left(1 - \frac{2}{\alpha} e^{-\alpha^2 / 2}\right),
\]
where $\alpha \geq 1$, $d \geq 4$, and $c$ is a positive constant.

Since $\mathbf{n}_i^T \mathbf{z}_i$ is normally distributed and symmetric about 0, the distribution of $\mathbf{n}^T \mathbf{z}$ is also symmetric. Therefore,
\[
\mathbb{P}\left(\left|\sum_{i=1}^{m} \mathbf{n}_i^T \mathbf{z}_i\right| \leq 2\alpha \sqrt{\frac{d}{d - 2}}\right) = \mathbb{P}\left(-2\alpha \sqrt{\frac{d}{d - 2}} \leq \sum_{i=1}^{m} \mathbf{n}_i^T \mathbf{z}_i \leq 2\alpha \sqrt{\frac{d}{d - 2}}\right).
\]
Using the symmetry of the distribution:
\[
\mathbb{P}\left(\left|\sum_{i=1}^{m} \mathbf{n}_i^T \mathbf{z}_i\right| \leq 2\alpha \sqrt{\frac{d}{d - 2}}\right) = 2\mathbb{P}\left(\sum_{i=1}^{m} \mathbf{n}_i^T \mathbf{z}_i \leq 2\alpha \sqrt{\frac{d}{d - 2}}\right) - 1.
\]
For large values of $2\alpha \sqrt{\frac{d}{d - 2}}$, the second term $-1$ can be ignored, giving us:
\[
\mathbb{P}\left(\left|\sum_{i=1}^{m} \mathbf{n}_i^T \mathbf{z}_i\right| \leq 2\alpha \sqrt{\frac{d}{d - 2}}\right) \approx \mathbb{P}\left(\sum_{i=1}^{m} \mathbf{n}_i^T \mathbf{z}_i \leq 2\alpha \sqrt{\frac{d}{d - 2}}\right).
\]

The probability can be further estimated using the product of individual probabilities due to the independence of $\mathbf{n}_i^T \mathbf{z}_i$:
\[
\mathbb{P}\left(\sum_{i=1}^{m} \mathbf{n}_i^T \mathbf{z}_i \leq 2\alpha \sqrt{\frac{d}{d - 2}}\right) \geq \prod_{i=1}^{m} \mathbb{P}\left(\mathbf{n}_i^T \mathbf{z}_i \leq 2\alpha \sqrt{\frac{d}{d - 2}}\right).
\]

Given the concentration inequality for each $\mathbf{n}_i^T \mathbf{z}_i$, we have:
\[
\prod_{i=1}^{m} \left[\left(1 - 3e^{-cd}\right)\left(1 - \frac{2}{\alpha} e^{-\alpha^2 / 2}\right)\right].
\]
For small values of $m$, this can be approximated as:
\[
\left[\left(1 - 3e^{-cd}\right)\left(1 - \frac{2}{\alpha} e^{-\alpha^2 / 2}\right)\right]^m.
\]

Combining all the steps, we arrive at the final concentration inequality:
\[
\mathbb{P}\left(\left|\sum_{i=1}^{m} \mathbf{n}_i^T \mathbf{z}_i\right| \leq 2\alpha \sqrt{\frac{d}{d - 2}}\right) \geq \left[\left(1 - 3e^{-cd}\right)\left(1 - \frac{2}{\alpha} e^{-\alpha^2 / 2}\right)\right]^m.
\]

\end{proof}

\subsection{Proof for Proposition 2}
\begin{proof}
    To prove the orthogonality of the extended vectors $\{\mathbf{n}_i^{\text{ext}}\}$, we calculate the inner product between any two distinct vectors $\mathbf{n}_i^{\text{ext}}$ and $\mathbf{n}_j^{\text{ext}}$ for $i \neq j$.

For each $i = 1, \ldots, m$, the vector $\mathbf{n}_i^{\text{ext}} \in \mathbb{R}^{md}$ can be written as
\[
\mathbf{n}_i^{\text{ext}} = (\underbrace{0, \ldots, 0}_{(i-1)d}, \mathbf{n}_i, \underbrace{0, \ldots, 0}_{(m-i)d}),
\]
where $\mathbf{n}_i$ is the $i$-th unit vector in $\mathbb{R}^d$, and the other entries are zero.

To find the inner product $(\mathbf{n}_i^{\text{ext}})^T \mathbf{n}_j^{\text{ext}}$ for $i \neq j$, we observe that the vectors $\mathbf{n}_i^{\text{ext}}$ and $\mathbf{n}_j^{\text{ext}}$ have non-zero components in different blocks. Specifically, $\mathbf{n}_i^{\text{ext}}$ has non-zero components only in the $i$-th block, while $\mathbf{n}_j^{\text{ext}}$ has non-zero components only in the $j$-th block. Since $i \neq j$, these blocks are disjoint. Therefore, for all $k$, either $(\mathbf{n}_i^{\text{ext}})_k = 0$ or $(\mathbf{n}_j^{\text{ext}})_k = 0$, or both. As a result, their inner product is zero:
\[
(\mathbf{n}_i^{\text{ext}})^T \mathbf{n}_j^{\text{ext}} = 0.
\]

Furthermore, for the same index $i$, the norm of the extended vector $\mathbf{n}_i^{\text{ext}}$ is preserved since it retains the norm of the original vector $\mathbf{n}_i$:
\[
\|\mathbf{n}_i^{\text{ext}}\|^2 = (\mathbf{n}_i^{\text{ext}})^T \mathbf{n}_i^{\text{ext}} = \mathbf{n}_i^T \mathbf{n}_i = 1.
\]

Thus, the extended vectors $\{\mathbf{n}_i^{\text{ext}}\}_{i=1}^{m}$ are orthogonal in $\mathbb{R}^{md}$, with each vector having unit norm. This completes the proof.
\end{proof}

\section{Experiment Details}
\subsection{Implementation Details}
\label{sec imple details}
\textbf{Implementation Detail}. We have included three diffusion transformers for our experiments: Stable Diffusion V3~\cite{esser2024scaling}, Stable Diffusion V3.5, and Flux. They are the most popular and state-of-the-art open-source transformer-based Text-to-Image (T2I) diffusion model. % which was pre-trained on ImageNet~\cite{} and CC12M~\cite{} datasets. 
We follow the implementation of Stable Diffusion V3 of the Huggingface Diffusers library~\cite{von-platen-etal-2022-diffusers}, and use the provided checkpoint with the model card "stabilityai/stable-diffusion-3-medium-diffusers". Similarly, for Stable Diffusion V3.5, we utilize use the provided checkpoint with the model card "stabilityai/stabilityai/stable-diffusion-3.5-medium". For qualitative results, we set the classifier-free-guidence (CFG) scale to 7.0. For the image editing task, we set the CFG scale to 7.5 and the total sampling steps to 50, we first forward the image to $75\%$ of the total timesteps, and then conduct the reverse process. For our UNet-based T2I model, we utilize Stable Diffusion V2.1 from Huggingface Diffusers~\cite{von-platen-etal-2022-diffusers}, which is the most widely used UNet-based T2I diffusion model. We set the hyperparameters the same as those mentioned above. During our experiments, we kept diffusion models frozen for zero-shot image-editing task. For the implementation of our method, we utilize the GPT-4~\cite{achiam2023gpt} to extract text description of given images. Details of the prompts input to the multi-modal LLM to obtain the text descriptions can be found in the appendix.

\textbf{Baseline Editing Methods.} Since text-guided zero-shot precise image editing is a novel task, we adapted several existing image editing methods with text-to-image diffusion models. Specifically, we utilize DiffEdit~\cite{couairon2022diffedit}, Pix2Pix~\cite{brooks2023instructpix2pix}, and MasaCtrl~\cite{cao2023masactrl} as our baselines. Most of these methods rely on UNet-specific network structures or embeddings, so we implemented their backbones using a UNet-based diffusion model. To enable precise image editing, we applied similar text manipulation techniques to the text embeddings that guide the reverse process of image editing.

\textbf{Qualitative Evaluation.} We conducsed a comprehensive qualitative assessment of EIM using a diverse array of real images spanning multiple domains. Our evaluation employed simple text prompts to describe various editing categories, including but not limited to style, appearance, shape, texture, color, action, lighting, and quantity. These edits were applied to a wide range of objects, such as humans, animals, landscapes, vehicles, food, art, and moving objects elements. To demonstrate the necessity of image-side manipulation, we compared  EIM 's performance with text-only embedding space manipulations. Table \ref{tab:table2} presents a checklist of the qualitatively investigated image editing tasks along with illustrative examples. For our base images, we generated high-resolution samples using Stable Diffusion 3. We then applied EIM with 3 different editing degrees to showcase precise editing capabilities and the controllability of the editing degree. 

Additionally, we compared the precise editing results of our EIM method with those of the baseline models. We adapted the baselines to the same experimental pipeline for precise image editing. Like  EIM, these methods were applied with three different editing degrees to demonstrate their capabilities. The comparison highlights EIM's superior ability to achieve precise edits and control the degree of modification.

\textbf{Benchmark Construction and Quantitative Evaluation}
Despite recent advancements in diffusion-based image editing, there has been a notable absence of well-labeled, objective, and precise evaluation benchmarks, particularly for zero-shot editing. While \cite{kawar2023imagic} introduced TEDBench, a 100-image benchmark for precise image editing, its reliance on human evaluation makes it time-consuming and costly to replicate across different models. Similarly, \cite{ju2023direct} developed PIEBench with both human and machine evaluations, but it lacks a comprehensive assessment of precise editing, especially regarding the degree of edit features.
To address these limitations and quantify the effectiveness of our EIM framework, we introduce ZOPIE (Zero-shot Open-source precise Image Editing benchmark). ZOPIE is the first diffusion-based image editing benchmark that incorporates both human subjective and automatic objective machine evaluations, assessing image quality, controllability of precise edits, image-text consistency, background preservation, and semantic disentanglement.
ZOPIE comprises 576 images evenly distributed across 8 editing types and 8 object categories (humans, animals, landscapes, vehicles, food, art, moving objects, and sci-fi elements). Each benchmark element includes source prompt, edit prompt, editing feature, object class, object region of interest mask, and background mask annotations.
For human subjective evaluation, we conducted an Amazon Mechanical Turk user study following the standard Two-Alternative Forced Choice (2AFC) protocol. We gradually increased the editing strength of EIM and other baseline methods, producing three consecutive image edits. Four participants with diverse backgrounds were presented with our edits and baseline edits, and asked to select the superior set of images based on editing quality, background preservation, and edit-prompt-image consistency. In total, 20,736 results were collected.
For automatic and objective evaluation, we employed six metrics across four aspects:
\begin{itemize}
    
\item Background preservation: PSNR, LPIPS, and SSIM
\item Text-image consistency: CLIPScore
\item precise editing effectiveness: MLLM-VQA score (Multimodal LLM-based Visual Question Answering score) This metric uniquely leverages advanced multimodal language models to assess the effectiveness of gradual precise editing. By posing targeted questions about consecutive edits to a large language model, we quantify the perceptibility and smoothness of incremental changes in the edited images.
%\item Semantic disentanglement: Semantic Disentanglement Metric (SDE) This metric quantifies the ability to edit specific features in the latent space without affecting others. It provides a numerical measure of how well the editing process preserves unrelated semantic attributes while modifying the target feature, addressing a critical aspect of precise editing that previous benchmarks overlooked.
\end{itemize}
The CLIP score was calculated between the edit prompt and the edited images. For the MLLM-VQA score, we input pairs of images with consecutive editing strengths into GPT-4 and posed the question: "Answer only in yes or no, does the second image reflect a gradual change of [edit feature] compared to the first image?" We ran this process five times per sample and calculated the percentage of "yes" responses across all runs and samples.

\textbf{Evaluation on semantic disentanglement degree.} We employ the SDE metric proposed in Sec.~\ref{SDE} to quantitatively assess the disentanglement degree of text-to-image diffusion models. Specifically, we calculate the SDE for both the UNet-based diffusion model and the diffusion transformer using the CeleBA~\cite{liu2015deep} dataset. CeleBA is a widely used benchmark for evaluating the generative capabilities of diffusion models~\cite{rombach2022high,zhang2024distributionally}, associates each image with an attribute vector, making it particularly suitable for measuring the disentanglement degree with the SDE metric. We randomly sample 2000 samples from the whole dataset, and test the disentanglement degree of the models based on age, gender, expression, hair, eyeglasses, and hat attributes.

\subsection{Multi-Modal-LLM prompts}
\label{sec mmllm}
As detailed in Sec.~\ref{sec  EIS }, our EIM method leverages a multi-modal LLM, specifically GPT-4, to generate text descriptions for both the source image and the desired edited image. An example of the input and output is shown in Fig.~\ref{fig:appendix example}:

\begin{figure}
    \centering
    \includegraphics[width=1\linewidth]{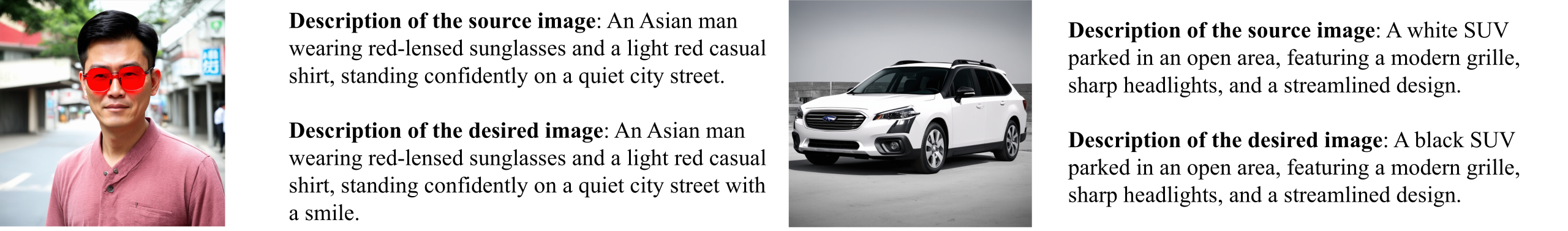}
    \caption{Example of the text descriptions of the given source image.}
    \label{fig:appendix example}
\end{figure}

\begin{figure}
    \centering
    \includegraphics[width=1\linewidth]{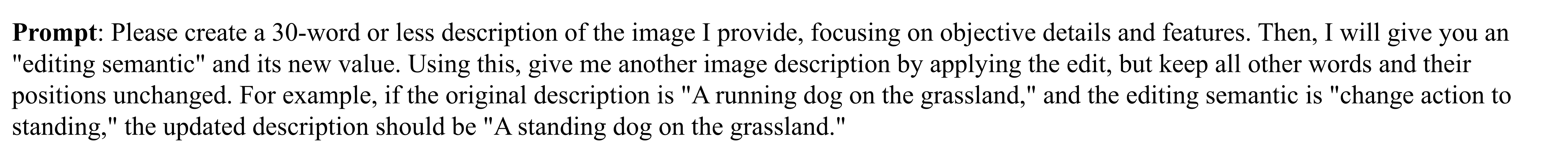}
    \caption{Instructions input to multi-modal LLM before the Extract stage in our EIM method.}
    \label{fig:prompt}
\end{figure}

To ensure that the multi-modal LLM captures the fine details of the source image and maintains consistency in task-irrelevant text tokens between the descriptions of the source image and the desired edited image, we first guide the description process with specific prompts before inputting the source image, as shown in Fig.~\ref{fig:prompt}.

\subsection{Hyper-parameters}
In our EIM method, we empirically set the editing ratio between 0.2 and 1.0. When working with high-resolution images (e.g., 768x768) using the stable diffusion 3 model, we find that the effective threshold for the editing ratio is approximately 20 to 50. Exceeding this threshold significantly reduces image quality. During the sampling stage, we set the step size between 0.05 and 0.5, and the regularization degree between 0.2 and 0.5 to balance the precise editing and image quality.

\subsection{Probing Analysis}
Following~\cite{clark2019does,liu2019linguistic}, we conduct a probing analysis to investigate whether the category information of a semantic in UNet-based diffusion models and diffusion transformers is embedded within representations of other semantics. Inspired by~\cite{liu2024towards}, we perform a probing analysis on attention maps corresponding to the targeted semantics. We generate images using the caption "a $<$color$>$ $<$object$>$" and then analyze the attention maps corresponding to each token within the intermediate layers. We first train classifiers on the attention maps associated with "$<$color$>$," and then test these classifiers on the attention maps for "$<$object$>$." If these classifiers can accurately identify color information within the "$<$object$>$" attention maps, this suggests that the attention maps may include category information related to "$<$color$>$." In our experiments, we used the text prompt "a $<$color$>$ car" and extracted the attention maps for the three words in each layer. We trained binary classifiers on the attention map for "$<$color$>$" and tested them on the attention maps for the other words. If the classifiers can determine whether a specific color is present, the attention map may contain category information about color. The closer the accuracy is to 0.5, the less category information is embedded in the attention map. For implementation, we consider 3 colors, and for each color, we generate 1000 images to obtain the attention maps, and train a linear layer to do the linear probing analysis.

\section{Additional Analysis Experiments}

\subsection{Extension to Multi-Attribute Editing}
Our EIM method can be extended to multi-attribute editing scenarios. Consider a semantic $s\in \mathcal{S}$ that consists of $m$ modular semantics $s_i \in \mathcal{S}_i, i\in [1,...,m]$. We extend it to the entire space to get $\mathbf{n}_i^{\text{ext}}\in \mathbb{R}^{md}$. As discussed in Appendix~\ref{sec:multi-attr edit}, the extended editing directions of different semantics are orthogonal to each other. Therefore, given editing directions \( \mathbf{n}_1, \mathbf{n}_2, \dots, \mathbf{n}_m \in \mathbb{R}^d \) for semantics \( s_1, s_2, \dots, s_m \) and corresponding editing degrees \( \alpha_1, \alpha_2, \dots, \alpha_m \), each \( \alpha_i \cdot \mathbf{n}_i \) can be applied individually. These are then combined into an extended editing direction \( \mathbf{n}^{\text{ext}} = (\alpha_1 \cdot \mathbf{n}_1, \alpha_2 \cdot \mathbf{n}_2, \dots, 0_d) \in \mathbb{R}^{md} \).

Specifically, for single-semantic manipulation on the text side, we compute the editing direction $n$ by $c_1-c_0$, where $c_1,c_0$ is the embedding of extracted text description of the source image and desired image. Therefore, we used the following formula for text-side manipulation:
\[
\mathbf{c} = \mathbf{c}_0 + \alpha (\mathbf{c}_1 - \mathbf{c}_0)
\]
where $\mathbf{c}_0$ is the original embedding, $\mathbf{c}_1$ is the target embedding, and $\alpha$ is the editing strength.

\begin{figure*}[t!] 
    \centering
    
    \includegraphics[width=1\textwidth]{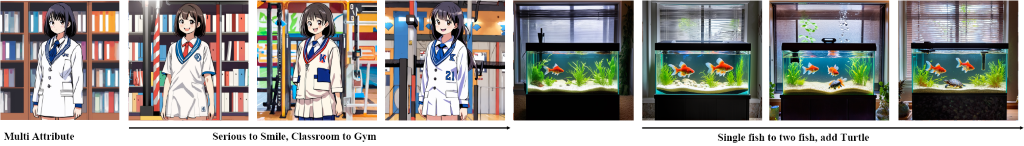} % Adjust the path and width as needed
    %\vspace{-10pt}
    \caption{Extension to multi-attribute precise editing. EIM method has successfully edited multiple attributes simultaneously.
    }
    %\vspace{-10pt}
\label{multi-attr fig}
\end{figure*}

For multi-attribute editing, we can extend this to:
\[
\mathbf{C} = \mathbf{C}_0 + \Lambda (\mathbf{C}_1 - \mathbf{C}_0)
\]
Here, $\Lambda$ is a diagonal matrix indicating the editing degree of multiple attributes. This formulation leverages the decomposability of the disentangled semantic representation space, allowing us to edit multiple attributes simultaneously without interference. As shown in Fig.~\ref{multi-attr fig}, our EIM has successfully achieved precise multi-attribute editing.

\begin{figure}
    \centering
    \includegraphics[width=1\linewidth]{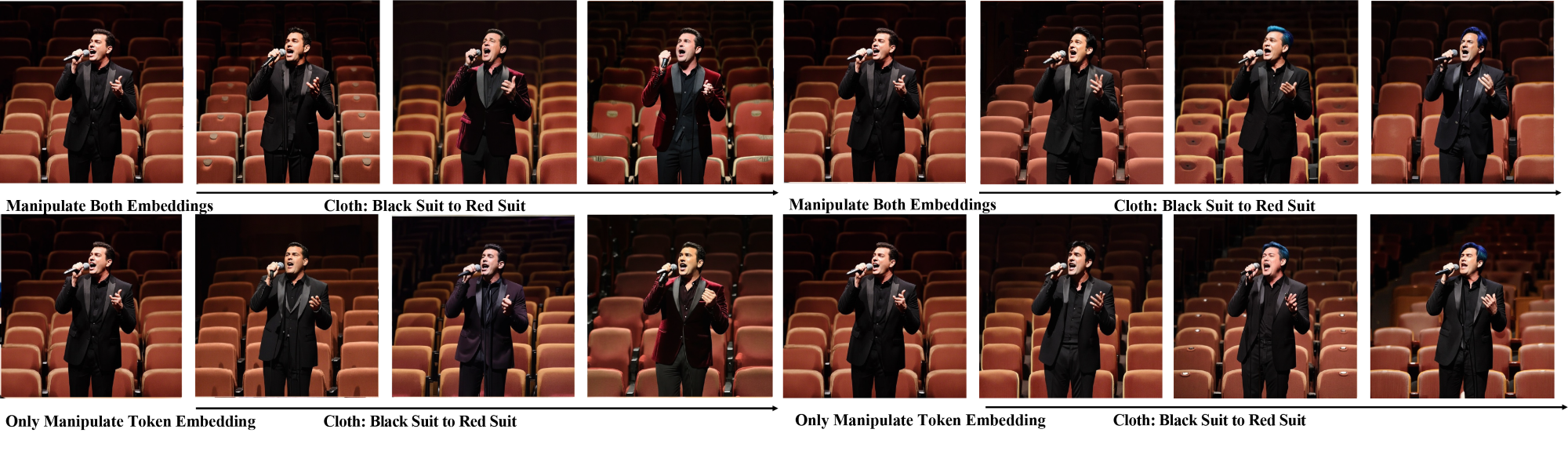}
    \caption{Manipulate both text embedding and the pooled text embedding provides more precise image editing results.}
    \label{fig:ablation_pool}
\end{figure}

\textbf{Manipulating pooled text embedding can also boost the precise editing performance.} Figure~\ref{fig:ablation_pool} illustrates the impact of manipulating different text embeddings in text-guided image editing. When both text embedding and the pooled text embeddings are manipulated, the editing results are more precise and controlled, as demonstrated in the transition from a black suit to a red suit. In contrast, editing only the token embeddings results in less precise modifications, often failing to achieve the intended effect comprehensively. This highlights the importance of incorporating pooled text embeddings to achieve more refined and accurate edits.

\section{Supplementary Qualitative Results}
\label{sec supplementary}
%Here we provide more qualitative results in Fig.~\ref{} and Fig~\ref{}, in supplementary for Sec.~\ref{}.

Also, we have supplemented our study with additional qualitative comparisons against baseline approaches. EIM consistently outperforms these baseline methods in achieving more precise and precise image editing. As shown in Fig.~\ref{fig:supple baseline 1} and Fig.~\ref{fig:supple baseline 2}, baseline approaches often introduce unintended changes to irrelevant semantics, which disrupt the overall coherence of the image. In contrast, EIM demonstrates superior capability in making targeted, precise adjustments to specific semantics while maintaining the integrity of non-target attributes. This underscores  EIM's advanced ability to preserve image coherence and effectively execute the desired edits with greater precision.

\begin{figure*}[htp] 
    \centering
    
    \includegraphics[width=1\textwidth]{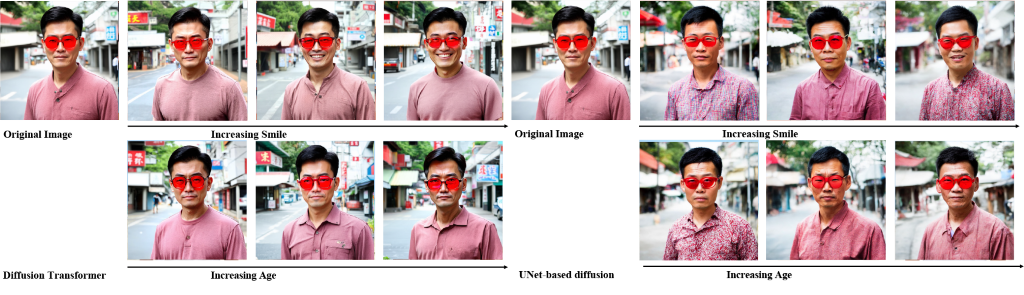} % Adjust the path and width as needed
    %\vspace{-10pt}
    \caption{Comparison of samples generated by diffusion transformer and traditional UNet-based diffusion model. Transformer-based diffusion models can achieve more precise and precise image editing on target semantics.
    }
    %\vspace{-10pt}
\label{sd2sd3}
\end{figure*}

\begin{figure}
    \centering
    \includegraphics[width=1\linewidth]{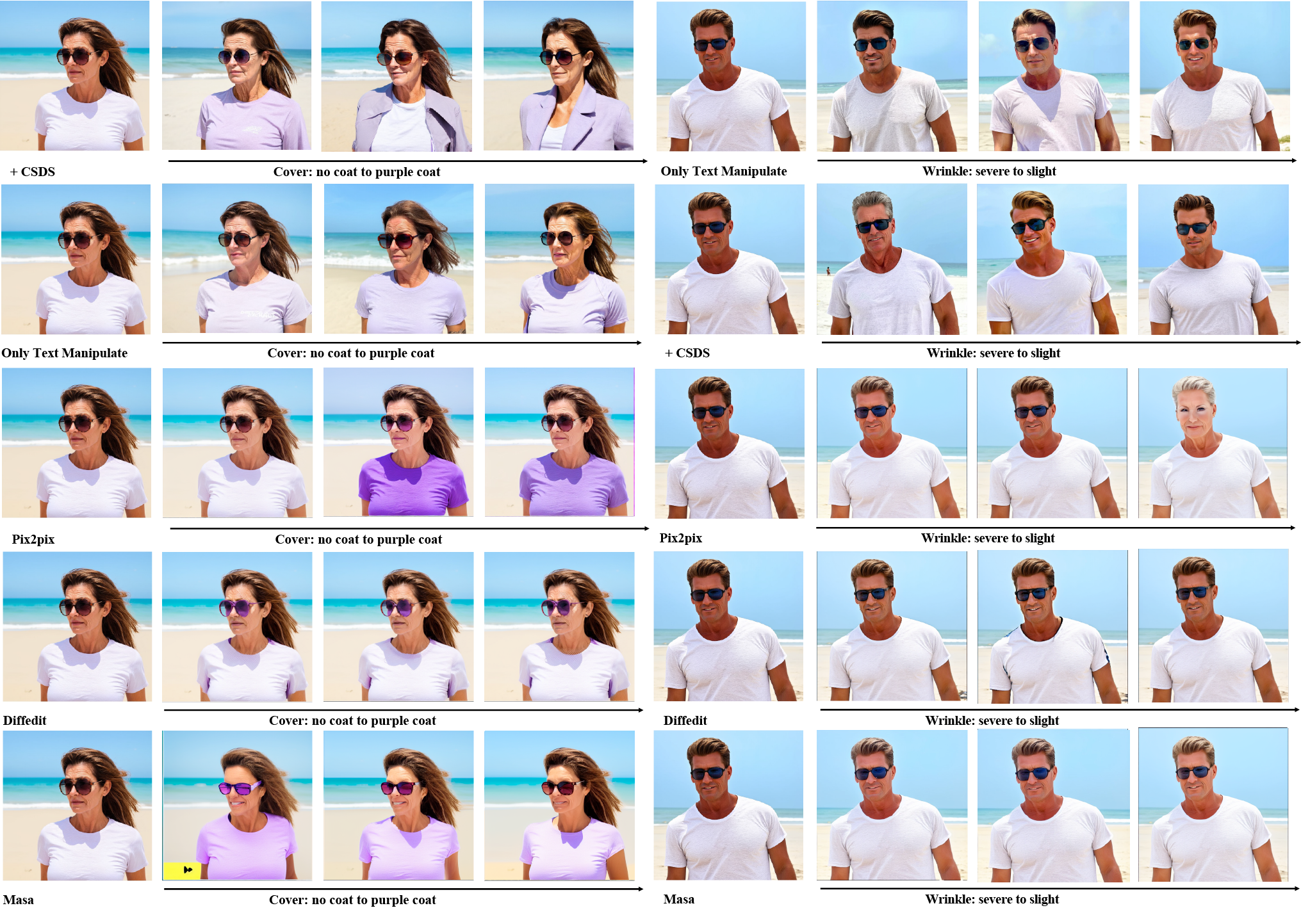}
    \caption{Comparisons between our EIM method and baseline methods, our method achieved the best precise object-based transformations from lion to tiger and pine tree to utility pole. EIM outperformed others in preserving image quality and achieving intended edits effectively.}
    \label{fig:supple baseline 1}
\end{figure}

\begin{figure}
    \centering
    \includegraphics[width=1\linewidth]{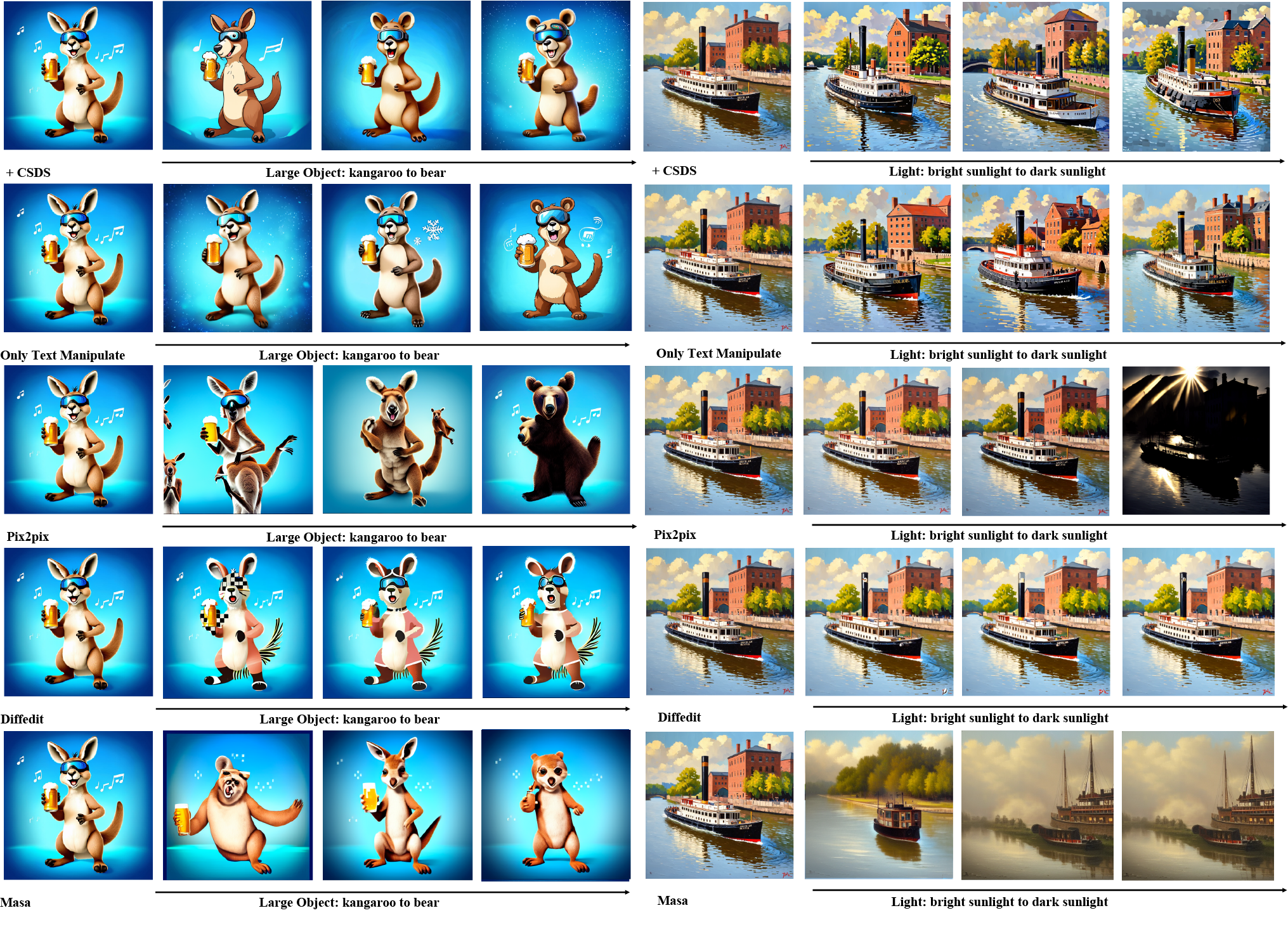}
    \caption{Comparisons between our EIM method and baseline methods, our method achieved the best precise object-based transformations from lion to tiger and pine tree to utility pole. EIM outperformed others in preserving image quality and achieving intended edits effectively.}
    \label{fig:supple baseline 2}
\end{figure}

\textbf{DiT possesses a better disentangled latent space than UNet diffusion.} As illustrated in Figure~\ref{sd2sd3}, the comparison between the diffusion transformer and the traditional UNet-based diffusion model clearly demonstrates the superior performance of the transformer-based approach in achieving precise and precise image editing. This advantage stems from the disentanglement property inherent in the diffusion transformer. In our experiments, we applied identical text-side manipulations to the input text embeddings for both models. When increasing the "smile" attribute, the transformer model consistently produces more precise and controllable modifications, whereas the UNet-based model introduces noticeable distortions in areas unrelated to the editing task, such as clothing, hair shape, and facial structure. Similarly, when editing the "age" attribute, the diffusion transformer preserves the integrity of other facial features, resulting in more realistic outcomes compared to the UNet-based model, which struggles to isolate changes to the target semantic. This comparison highlights the advanced capabilities of transformer-based diffusion models in managing complex, precise edits while maintaining overall image quality.

\begin{figure*}[h!] 
    \centering
    
    \includegraphics[width=1\textwidth]{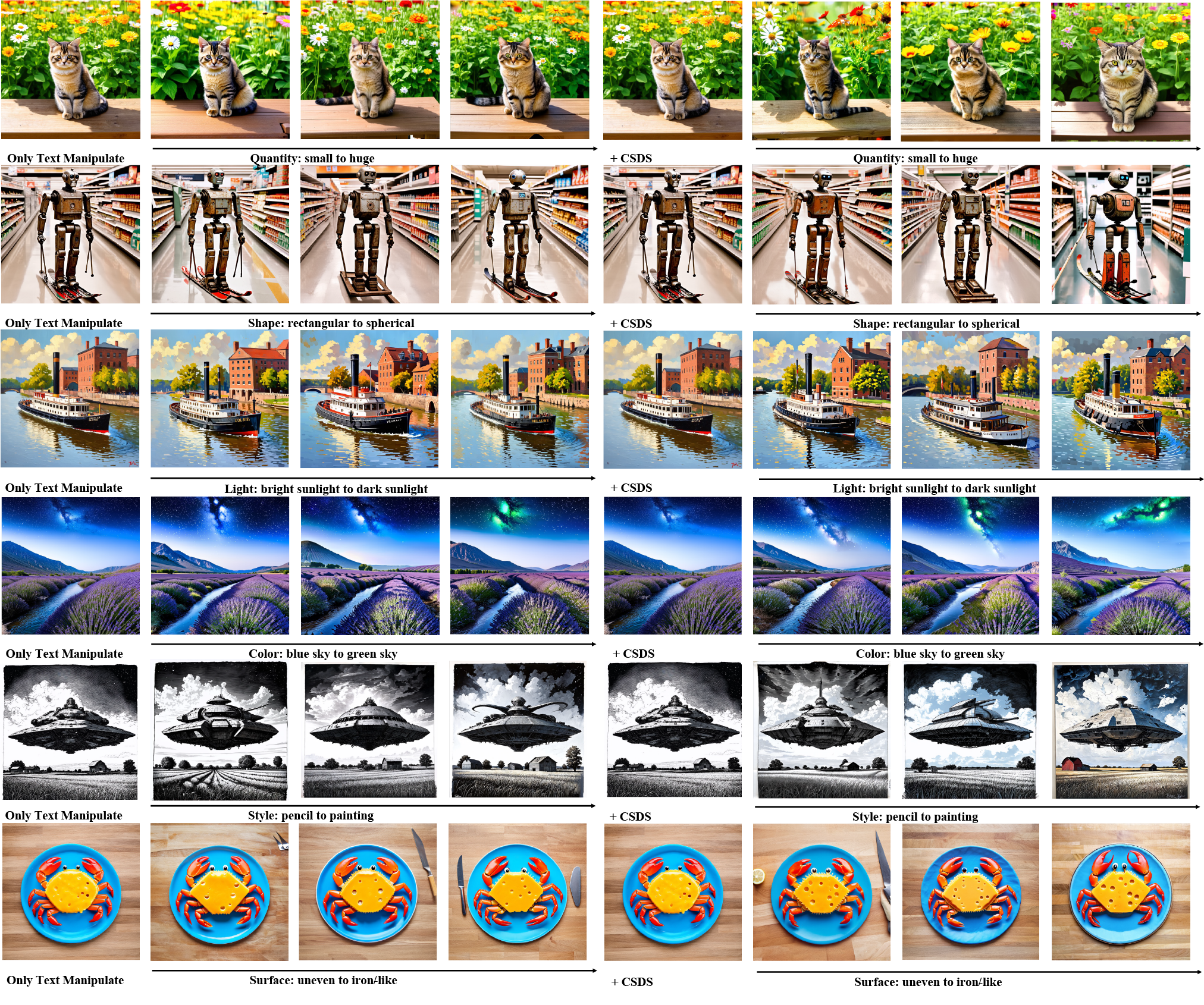} % Adjust the path and width as needed
    %\vspace{-10pt}
    \caption{Precise editing results of attribute-based and texture-based semantics. We utilize diffusion transformers to precisely modify the intensity of considered attributes.
    }
    %\vspace{-10pt}
\label{extra1}
\end{figure*}
\begin{figure*}[t!] 
    \centering
    
    \includegraphics[width=1\textwidth]{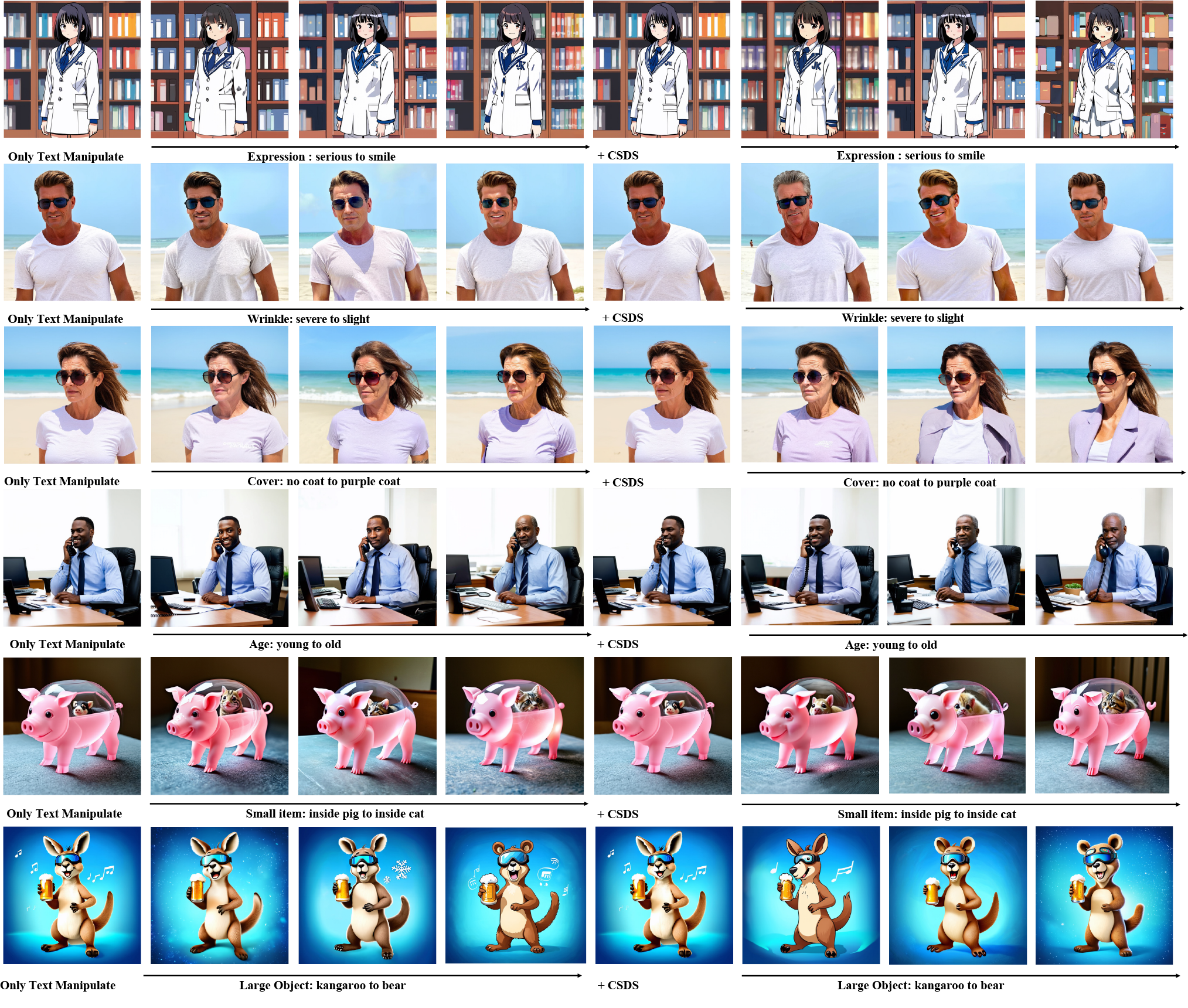} % Adjust the path and width as needed
    %\vspace{-10pt}
    \caption{Precise image editing results of person-based and object-based semantics. We utilize diffusion transformers to modify the intensity of considered attributes in a fine granularity.
    }
    %\vspace{-10pt}
\label{extra2}
\end{figure*}

\section{Supplementary Analysis Study}
\label{appendix analysis study}
\subsection{Detailed Results of Probing Analysis}
Detailed results of the probing analysis are shown below in Table~\ref{detail prob}:

\begin{table}[h]
\centering
\scalebox{0.68}{
\begin{tabular}{llll llll llll lll}
\toprule
Category & Backbone & Word & Layer 2 & Layer 4 & Layer 6 & Layer 8 & Layer 10 & Layer 12 & Layer 14 & Layer 16 & Layer 18 & Layer 20 & Avg \\
\midrule
green & Unet & "car" & 1.000 & 0.981 & 1.000 & 1.000 & 0.821 & 0.130 & 0.024 & 0.684 & 0.388 & 0.997 & 0.7025 \\
blue & Unet & "car" & 1.000 & 0.958 & 1.000 & 1.000 & 0.998 & 0.913 & 0.159 & 1.000 & 0.856 & 1.000 & 0.8884 \\
red & Unet & "car" & 0.996 & 1.000 & 0.957 & 0.009 & 0.919 & 0.737 & 0.043 & 1.000 & 0.852 & 1.000 & 0.7513 \\
green & transformer & "car" & 0.224 & 0.242 & 0.315 & 0.262 & 0.247 & 0.159 & 0.193 & 0.233 & 0.849 & 0.801 & 0.3525 \\
blue & transformer & "car" & 0.176 & 0.634 & 0.353 & 0.504 & 0.239 & 0.233 & 0.339 & 0.363 & 0.206 & 0.155 & 0.3202 \\
red & transformer & "car" & 0.805 & 0.515 & 0.410 & 0.367 & 0.311 & 0.328 & 0.297 & 0.287 & 0.510 & 0.390 & 0.4220 \\
green & Unet & "a" & 1.000 & 0.043 & 0.984 & 0.453 & 1.000 & 0.000 & 0.231 & 0.621 & 1.000 & 0.023 & 0.5355 \\
blue & Unet & "a" & 0.000 & 0.999 & 0.000 & 0.770 & 0.075 & 0.998 & 0.101 & 0.060 & 1.000 & 0.000 & 0.4003 \\
red & Unet & "a" & 1.000 & 0.002 & 0.999 & 0.000 & 0.000 & 0.842 & 0.005 & 0.997 & 0.000 & 0.018 & 0.3863 \\
green & transformer & "a" & 0.848 & 0.938 & 0.299 & 0.412 & 0.122 & 0.095 & 0.126 & 0.195 & 0.036 & 0.135 & 0.3206 \\
blue & transformer & "a" & 0.838 & 0.182 & 0.402 & 0.397 & 0.360 & 0.423 & 0.288 & 0.309 & 0.843 & 0.614 & 0.4656 \\
red & transformer & "a" & 0.091 & 0.305 & 0.360 & 0.221 & 0.337 & 0.254 & 0.375 & 0.340 & 0.165 & 0.497 & 0.2945 \\
\bottomrule
\end{tabular}
}
\caption{Layer-wise probing accuracies of attention maps of transformer-based and UNet-based diffusion models.}
\label{detail prob}
\end{table}

\section{Impact and Limitation}
\textbf{Impact.} This work represents the first in-depth study of the latent space in text-to-image diffusion transformers (T2I DiT), uncovering a disentangled semantic representation space. This discovery advances our understanding of diffusion transformers and generative models, highlighting their potential for precise, controllable image synthesis. Our EIM method, inspired by these findings, demonstrates effective precise control in image generation, making it a valuable tool in domains requiring high levels of controllability, trustworthiness, and interpretability. Beyond image synthesis, this approach can be extended to applications like video generation and content safety control. Furthermore, we introduce a novel metric and benchmark for evaluating disentanglement effects, providing a valuable reference for future research in this area.

\textbf{Limitations.} Our experiments on text-to-image diffusion transformers are mainly conducted on Stable Diffusion 3, which is the only transformer-based text-to-image diffusion model. Additionally, our focus was exclusively on the image synthesis task, though diffusion transformers like DiT have broader potential, such as serving as prior knowledge for various vision tasks. Finally, our analysis centers on pre-trained models, leaving unexplored the role of the training process in the formation of disentangled representations. Future work should empirically and theoretically investigate whether disentanglement is driven by DiT's architecture alone or also influenced by the training dynamics.

\end{document}